\documentclass[11pt, a4paper]{article}

\usepackage{amsmath,amsfonts,amssymb}
\usepackage{algorithm}
\usepackage{array}
\usepackage[caption=false,font=normalsize,labelfont=sf,textfont=sf]{subfig}
\usepackage{textcomp}
\usepackage{stfloats}
\usepackage{url}
\usepackage{verbatim}
\usepackage{graphicx}
\usepackage{cite}
\usepackage{color}
\usepackage{tablefootnote}

\newcommand{\XX}{\mathcal{X}}
\newcommand{\YY}{\mathcal{Y}}
\newcommand{\EE}{\mathcal{E}}
\newcommand{\QQ}{\mathcal{Q}}
\newcommand{\NN}{\mathcal{N}}
\newcommand{\UU}{\mathcal{U}}
\newcommand{\rr}{\mathbb{R}}
\newcommand{\st}{\mathop{\rm s.t.}\nolimits}
\newcommand{\eqdef}{\triangleq}
\newcommand{\QED}{\hfill$\Box$}

\newcommand{\ideal}{\textsf{ideal}}
\newcommand{\random}{\textsf{random}}
\newcommand{\greedyx}{\textsf{GS$_x$}}
\newcommand{\greedyxy}{\textsf{iGS}}
\newcommand{\QBC}{\textsf{QBC}}
\newcommand{\iRDM}{\textsf{iRDM}}

\newcommand{\beqar}{\begin{eqnarray}}
\newcommand{\eeqar}{\end{eqnarray}}
\newcommand{\beqarno}{\begin{eqnarray*}}
\newcommand{\eeqarno}{\end{eqnarray*}}
\newcommand{\ba}[1]{\begin{array}{#1}}
\newcommand{\ea}{\end{array}}

\usepackage{enumitem}
\renewcommand{\theenumi}{\arabic{enumi}}

\ifx\example\undefined
	\newtheorem{exampleenv}{Example}[section]
	\newcommand{\example}[2]{
	\begin{exampleenv}
	{
	\label{#1}
	{\textrm{#2}}
	}
	\hfill\QED
	\end{exampleenv}
	}
\fi

\begin{document}

\title{Active Learning for Regression by Inverse Distance Weighting}
\author{Alberto Bemporad
\thanks{The author is with the IMT School for Advanced Studies, Piazza San Francesco 19, Lucca, Italy. Email: \texttt{\scriptsize alberto.bemporad@imtlucca.it}.}}

\markboth{Submitted for publication}%
{A. Bemporad}

\maketitle

\begin{abstract}
This paper proposes an active learning (AL) algorithm to solve regression problems 
based on inverse-distance weighting functions for selecting the feature vectors to query. The algorithm has the following features: ($i$) supports both pool-based and population-based sampling; ($ii$) is not tailored to a particular class of predictors; ($iii$) can handle known and unknown constraints on the queryable feature vectors; and ($iv$) can run either sequentially, or in batch mode, depending on how often the predictor is retrained. 
The potentials of the method are shown in numerical tests on illustrative synthetic problems and real-world datasets. An implementation of the algorithm, that we call IDEAL 
(Inverse-Distance based Exploration for Active Learning), is available at \texttt{\url{http://cse.lab.imtlucca.it/~bemporad/ideal}}.
\end{abstract}


\section{Introduction}
Active learning (AL) strategies are used in
supervised learning to let the training algorithm ``ask questions''~\cite{Set12},
i.e., choose the feature vectors to query for the corresponding target value
during the training phase, usually based on the model 
learned so far. The main aim of AL is to possibly reduce the number of training samples required to train the model, or in other words, to get a model of the same prediction quality with a smaller dataset. This is particularly useful when knowing the target value associated with a given combination of features is an expensive operation,
for example, it may involve asking a human to ``label'' samples manually, running a costly and time-consuming laboratory experiment, or performing a complex computer simulation.

AL methods are usually categorized in \emph{query synthesis} (or \emph{population-based})
methods, in which the feature vector to query can be chosen arbitrarily,
\emph{pool-based} sampling methods, in which the vector can only be chosen within a given finite set (or ``pool'') of unlabeled values, and \emph{selective-sampling} methods, in which
vectors are proposed in a streaming flow and the AL algorithm can only decide online whether
to ask for the corresponding target or not~\cite{Set12}.

Several approaches to AL are available in the literature, see, e.g., the survey papers~\cite{Set12,SW10,FZL13,AKGHP14,KG20}. Most of the literature focuses on classification problems~\cite{AKGHP14,RM01}, although AL has been investigated also for regression~\cite{Mac92,CGJ96,SN09,DMB13,DB14,CZZ13,CZZ17,Wu19,WLH19,LJLFLW21}. We will describe in detail some of the most popular AL algorithms for regression in Section~\ref{sec:other-AL}. For a detailed and updated taxonomy of AL methods for classification, regression, and clustering we refer the reader to the recent survey paper~\cite{KG20}. 

As pointed out in~\cite{Wu19}, AL methods should collect data that are \emph{informative}, \emph{representative}, and \emph{diverse}, i.e., respectively, contain rich information for reducing modeling errors, cover portions of the feature vector space where the predictor is evaluated most frequently and in particular reject outliers, and explore such a space trying to avoid sampling the same regions too often. AL methods are often linked to a specific class of predictors, such as neural networks~\cite{Mac92} or mixtures of Gaussians and 
locally weighted regression~\cite{CGJ96}, or to a particular learning algorithm~\cite{SN09,DMB13,DB14,CZZ17}.
Moreover, AL methods can be computationally involved in the case optimal sampling is sought,
or in query-by-committee (QBC) methods~\cite{SOS92,RH95,BRK07} in which multiple predictors need to be retrained repeatedly to measure their disagreement.

In general, in AL the \emph{acquisition function} that is used to drive
the selection of the next sample has two components. The first is related to
the position of the feature vector within the feature-vector space and
is used for pure \emph{exploration} of that space. The second 
aims at the \emph{exploitation} of the target values acquired so far,
learning a model on the available feature vectors/target pairs and
using it for predicting target values. Such a model-based approach 
usually tries to estimate a form or another of target uncertainty, such as 
to locate feature vectors whose target is supposed to be farthest from the target
values already acquired~\cite{WLH19}, sample where a committee
of predictors mostly disagree~\cite{SOS92,RH95,BRK07}, or select the feature
vector that is expected to make the most change in the prediction function~\cite{CZZ17}.

AL is related to the problem of optimally designing experiments, whose origins
date back at least to the 30s~\cite{Fis35}, and has attracted an extensive literature for decades~\cite{BHH78}. Another problem related to AL is black-box derivative-free optimization~\cite{RS13} in which a surrogate of the objective function is learned
incrementally from a finite number of samples of it, such as in Bayesian optimization methods~\cite{SSWAD15}. Compared to solving a supervised learning problem, where the objective is to find a model that reproduces well the underlying process over the entire set of feature-vectors of interest, in black-box optimization the problem is somehow simpler, as the interest is limited to approximating the objective function well around one of its global minimizers.

\subsection{Contribution}
In this paper, we provide an AL framework for regression 
that is applicable to any prediction model, can address both pool-based and population-based settings, and is not computationally involved. 
By leveraging on ideas we previously investigated for global optimization based on surrogate functions~\cite{Bem20,BP21}, we propose an AL method in which the uncertainty associated with the currently available predictor and the exploration function used to sample the feature-vector space are characterized by \emph{inverse-distance weighting} (IDW) functions~\cite{Kus64,She68}.

The proposed algorithm that we call IDEAL (Inverse-Distance based Exploration for Active Learning) blends different requirements: informativeness, by sampling regions
of the feature-vector space
where model uncertainty is estimated to be large; representativeness, in the case of pool-based sampling, by possibly taking into account a density function similar to the one used in density-based spatial clustering approaches~\cite{EKSX96}; and diversity, using an IDW exploration term
that is higher far away from samples that have already been queried. The algorithm can also
handle constraints on the feature vectors that can be queried, that can either be known a priori
or even \emph{unknown}. The latter case covers the situation in which one discovers only after querying certain combinations of features that the corresponding target cannot be retrieved; for example,
because a specific physical experiment cannot be performed or a computer simulation does
not converge. Finally, the proposed algorithm can be run either sequentially, by retraining
the predictor after each successful query, or in batch mode, by retraining only
after querying a certain prescribed finite number of samples.

IDEAL belongs to the class of model-based AL methods for regression, in that the
prediction model learned on the currently available samples is used in combination
with IDW terms to quantify model uncertainty and look for samples
that are expected to provide maximum informativeness. 
Moreover, the latter only requires the currently learned predictor,
contrarily to QBC methods that require instead training multiple 
predictors, and does not involve complex computations required 
by optimal sampling methods, which makes them only applicable
to relatively simple prediction models.

Similarly to the greedy sampling method proposed in~\cite{WLH19},
which combines diversity in the feature vector and (predicted) target spaces,
IDEAL combines the informativeness measure mentioned above with the diversity 
quantification in the feature-vector space provided by IDW terms only.
As we will show in several numerical examples, such a combination of model-based
uncertainty characterization and feature-vector diversity is beneficial with respect
to uncertainty characterization only, as in QBC methods~\cite{RH95,BRK07}, and 
input diversity only, as in the greedy method~\cite[Algorithm 1]{YK10}
and the improved representativeness-diversity maximization method~\cite{LJLFLW21}.

The paper is organized as follows. After formulating the AL problem in Section~\ref{sec:AL-problem},
we describe the proposed algorithm in Section~\ref{sec:IDEAL}. Numerical tests on synthetic
and real-world regression 
problems are reported in Section~\ref{sec:results}
and some conclusions are drawn in Section~\ref{sec:conclusion}.

A Python implementation of IDEAL, and of other passive and active learning methods
we have compared with, is available at \texttt{\url{http://cse.lab.imtlucca.it/~bemporad/ideal}}.

\section{Active learning problem}
\label{sec:AL-problem}
We consider a process $y:\XX\to\YY$ generating data $y_k=y(x_k)$,
where $\XX\subseteq\rr^n$ is the set of feature vectors, $x_k\in\XX$, and
$\YY\subseteq\rr^m$ the set of corresponding targets $y_k$, $y_k\in\YY$.
As the process $y$ is unknown, we wish to find a predictor $\hat y:\XX\to\YY$
solving the supervised learning problem 
\begin{equation}
    \min_{\hat y}\int_{\bar\XX\cap\XX}\ell(y(x),\hat y(x),x)dx    
\label{eq:supervised_learning}
\end{equation}
where $\ell:\YY\times\YY\times\XX\to\rr$ is a loss function,
for instance $\ell(y(x),\hat y(x),x)=\|y(x)-\hat y(x)\|_2^2$,
and $\bar\XX\subseteq\rr^n$ is a bounded set of feature vectors $x$
of interest, i.e., for which we want to obtain a 
good approximation $\hat y(x)$ of $y(x)$. While the set $\bar\XX$
is known, for example, it may be defined by the set of inequality constraints
\[
    \bar\XX=\{x:\rr^n:\ g_i(x)\leq 0,\ i=1,\ldots,n_c\}
\]
$g_i:\rr^n\to\rr$, the set $\XX$ for which $y(x)$ is defined could be \emph{unknown}, 
as we might not be able to know a priori whether for a 
given $x\in\bar\XX$ its corresponding target $y(x)$ can be obtained. 
For example, all features $x_i$ of interest may take any value 
between -10 and 10, i.e., $\bar\XX=\{x:\rr^n:\ |x_i|\leq 10,\ i=1,\ldots,n\}$
and $y(x)=\log(x)$, which is only defined for $x>0$. In this case,
$y(x)$ cannot be queried when $x\leq 0$, i.e., $\XX=\{x: x_i>0,\ i=1,\ldots,n\}$
and we are in the presence of the unknown constraint $x\in\XX$.

In practical real-world applications, unknown constraints may arise when
evaluating $y(x)$ may require running a complex experiment
or computer simulation, and this could not be completed for various reasons
for the particular parameter settings defined by $x$.
In such cases, characterizing the shape of $\XX$, if of interest, would be a binary
classification problem itself that is amenable for active learning.
Note that in the case of multiple targets ($m>1$), we could
generalize the setting by assuming that each process component
$[y]_i:\XX_i\to\YY_i$, $i=1,\ldots,m$. However, for simplicity of notation,
we assume here that $\XX=\cap_{i=1}^m\XX_i$, i.e.,
that either the entire output vector $y(x)$ is defined
or it is entirely undefined at a given $x$.

Special cases of~\eqref{eq:supervised_learning} are (multivariate)
regression problems ($\YY=\rr^m$) and classification problems
($\YY=\{0,1\}^m$). We assume that possible discrete features have been one-hot encoded,
and that hence in general $\XX\subseteq\{0,1\}^{n_b}\times\rr^{n_n}$, where $n_b$
and $n_n$ are the number of binary and numeric features, respectively,
$n=n_b+n_n$,
and that the loss $\ell$ contains impulsive terms (Dirac delta terms) so that~\eqref{eq:supervised_learning} can be rewritten as
\begin{equation}
    \min_{\hat y}\sum_{x_b\in\XX_b\cap \bar\XX_b}\int_{\XX_c\cap\bar\XX_c}
\ell(y(x),\hat y(x),x_c,x_b)dx_c
\label{eq:supervised_learning-gen}
\end{equation}
where $x_b$ denotes the subvector of binary components of the feature vector $x$,
$\XX_b$ ($\bar\XX_b$) the corresponding set of their admissible combinations
(of interest), and $\XX_c$ ($\bar \XX_c$) the set of admissible subvectors $x_c$ of numeric features (of interest).

In order to address problem~\eqref{eq:supervised_learning}, we will solve
its empirical approximation
\begin{equation}
    \min_{\hat y}\frac{1}{N}\sum_{k=1}^N\ell(y_k,\hat y(x_k),x_k)
\label{eq:supervised_learning-emp}
\end{equation}
where $D_N\eqdef\{(x_k,y_k)\}_{k=1}^N$ is a training dataset, 
with $y_k=y(x_k)$ for some unknown function $y$\footnote{
Although function $y$ is rather arbitrary, the formulation could be extended to explicitly include a noise term $\eta_k\in\rr^{n_\eta}$, so that $y_k=y(x_k,\eta_k)$
is available rather than $y(x_k)$. This would allow modeling non-reproducible
queries, i.e., $y_k\neq y_j$ for $x_k=x_j$, $k\neq j$.}.

In (supervised) \emph{passive} learning the training dataset $D_N$
is given, where clearly $x_k\in\XX$ for all $k=1,\ldots,N$, as
the corresponding targets $y_k$ have been acquired.
Instead, in \emph{active} learning we are free to select the training
vectors $x_k$ to \emph{query}, i.e., for which we want to get the corresponding target value $y_k$, if it is defined, or a declaration that $x_k\not\in\XX$.
We have a \emph{pool-based} AL problem when $x_k$ can only be selected from a pool
\begin{equation}
    \XX_P=\{\bar x_j\}_{j=1}^{M}
\label{eq:XX_P}
\end{equation}
 of samples, $M\geq N$, with $\XX_P\subseteq\bar\XX$, or a \emph{population-based} AL problem when $x_k$ can be chosen freely within the given bounded set $\XX_P=\bar\XX$.

\section{Active learning algorithm}
\label{sec:IDEAL}
Let $[x_{\rm min},x_{\rm max}]\subset\rr^n$ be the smallest hyper-box containing 
the feature vectors we are allowed to sample, i.e.,
\begin{subequations}
\begin{equation}
    [x_{\rm min}]_i\eqdef\min_{x\in\XX_P} [x]_i,\  [x_{\rm max}]_i\eqdef\max_{x\in\XX_P} [x]_i    
\label{eq:scaling-1}
\end{equation}
which in case of pool-based AL~\eqref{eq:XX_P} is equivalent to setting
\begin{equation}
    [x_{\rm min}]_i\eqdef\min_{j=1,\ldots,M} [\bar x]_j,\  [x_{\rm max}]_i\eqdef\max_{j=1,\ldots,M} [\bar x]_j    
\label{eq:scaling-2}
\end{equation}
In order to be immune to different scaling of the individual features, when querying samples we consider the scaling function $\sigma:\rr^n\to\rr^n$ defined as
\begin{equation}
    \sigma_i(x) \eqdef \frac{2}{[x_{\rm max}]_i-[x_{\rm min}]_i}\left(x_i-\frac{[x_{\rm max}]_i+[x_{\rm min}]_i}{2}\right),\ i=1,\ldots,n    
\label{eq:sigma}
\end{equation}
\label{eq:scaling}%
\end{subequations}
where clearly $\sigma(x)\in[-1,1]^n$ for all $x\in[x_{\rm min},x_{\rm max}]$. 

Let $N_{\rm max}$ be the total budget of queries we have available to perform the AL task.
During AL, we collect in the set\footnote{In case of multiple targets $m>1$ and
different feasible sets $\XX_i$, i.e., 
$[y]_i:\XX_i\to\YY_i$, one could define a separate set $\QQ_i$ for each
target $i=1,\ldots,m$, with $k\in\QQ_i$ if and only if $x_k\in\XX_i$} 
$\QQ\subseteq\{1,\ldots,N_{\rm max}\}$ the indices of the samples $x_k$ that have been selected 
and for which the corresponding target could be acquired, i.e., $k\in\QQ$ if and only if
$x_k\in\XX$. 
Moreover, in the case of pool-based sampling, we keep track of the indices of samples already extracted and queried from the pool $\XX_P$ in the set $\EE\subseteq\{1,\ldots,M\}$, to avoid possibly querying them again.

\subsection{Initialization}
\label{sec:init}
Before fitting any prediction model, as commonly done in most AL approaches we must first select $N_i$ samples $x_1,\ldots,x_{N_i}\in\bar\XX\cap\XX$. As also mentioned in~\cite[Section 4.3]{LJLFLW21}, unsupervised AL (i.e., AL that only selects samples based on their position
within the feature-vector space, without querying targets) can be superior to model-based AL
when the number of samples is small, due to the possibly high inaccuracy of $\hat y$ (and of the estimate of its uncertainty) when trained on a small set of samples. In fact, without first selecting
$x_1,\ldots,x_{N_i}$ in an unsupervised way, the first trained predictors $\hat y$ could drive the search quite inefficiently, especially when the exploration term is not dominant, leading to collecting weakly informative samples.
As a consequence, model-based criteria would remain quite inexact, leading
to further collect not-so-relevant samples, with consequent performances possibly even worse
than just randomly sampling $\bar\XX$.

In the case of population-based AL, we use Latin Hypercube Sampling (LHS)~\cite{MBC79} 
on the hyper-box $[x_{\rm min},x_{\rm max}]$;
in the case of pool-based AL, we run instead the K-means algorithm~\cite{Llo57} on
the pool $\XX^\sigma_P\eqdef \sigma(\XX_P)$ of scaled samples with $K=N_i$ and pickup the $N_i$ different vectors $\sigma(\bar x_1),\ldots,\bar \sigma(\bar x_{N_i})\in\XX^\sigma_P$ that are closest to the centroids obtained by K-means
in terms of Euclidean distance (cf.~\cite{Wu19}). As some vectors may be infeasible ($\bar x_k\not\in\bar\XX$)
or cannot be queried ($\bar x_k\not\in \XX$), similarly to the LHS algorithm with constraints described in~\cite[Algorithm~2]{Bem20} the vectors $\bar x_k\not\in\bar\XX\cap\XX$ are discarded, and the 
above procedure is repeated until a set of $N_i$ pairs $(\bar x_k,\bar y_k)$ is collected. 

We denote by $N_{\rm init}$, $N_{\rm init}\geq N_i$, the total number of samples queried during 
the initialization phase and by $\{(x_i,y_i)\}$, $i=1,\ldots,N_i$ the resulting set of collected samples. Note that in the case $\XX\subset\bar\XX$, $N_{\rm init}>N_i$ queries might be required to get $N_i$ pairs $(x_k,y_k)$, as samples $x\in\bar\XX\setminus\XX$ might be encountered for which $y(x)$
is not defined. In this case, $N_{\rm init}\in\QQ$, as the initialization phase terminates as 
long as $N_i$ pairs have been successfully collected. Note also that in case $N_i$
valid samples cannot be retrieved at initialization within the total budget $N_{\rm max}$ of queries we have available, the AL task cannot proceed further. In the case of absence or
irrelevance of unknown constraints ($\bar\XX\subseteq\XX$), we always have $N_{\rm init}=N_i$.

\subsection{Query-point selection}
Assume that we have collected $N$ samples $x_k$ and, $\forall k\in\QQ$, the corresponding target
values $y_k$, and that we have fit a predictor
$\hat y(x)$ on them by solving the supervised learning problem
as in~\eqref{eq:supervised_learning-emp}
\begin{equation}
    \hat y=\arg\min_{\hat y}\sum_{k\in\QQ} \ell(y_k,\hat y(x_k),x_k)
\label{eq:yhat}
\end{equation}

Then, we need to define a criterion to select the remaining $N_{\rm max}-N_{\rm init}$ samples $x_k$ to query. In this paper, we will select the next sample $x_{N+1}$ to query
by maximizing an \emph{acquisition function} $a:\rr^n\to[0,+\infty)$ that we will introduce in the sequel
\begin{equation}
    x_{N+1}=\arg\max_{x\in\XX_P} a(x)    
\label{eq:acquisition}
\end{equation}
retrain $\hat y$, update the
acquisition function $a$, increase $N$, and so on, until $N=N_{\rm max}$, i.e., the total available
budget for queries is exhausted. In case $y(x_{N+1})$ is not defined because $x_{N+1}\not\in\XX$, we clearly do not need to retrain $\hat y$.
The approach can be extended easily to batch-mode active learning by retraining $\hat y$ only after $T$ new queries have been performed, $T>1$.

To define the acquisition function $a$, we want to use an empirical estimation of the uncertainty $s_i(x)$, $s_i:\rr^n\to[0,+\infty)$, associated with each component $i$ of the prediction $\hat y(x)$,
$i=1,\ldots,m$, that we define here as we proposed in~\cite{Bem20} to promote exploration in global optimization using surrogate functions. 

Given a set $\{x_k\}_{k=1}^N$ of vectors of $\rr^n$, we consider the squared (scaled) Euclidean distance 
function $d^2:\rr^{n}\times \rr^n\to\rr$
\begin{equation}
    d^2(x,x_k)=\|\sigma(x_k)-\sigma(x))\|_2^2,\ i=1,\ldots,N
\label{eq:distance}
\end{equation}
In standard IDW functions~\cite{She68}, the weight functions 
$w_k:\rr^n\setminus\{x_k\}\to\rr$ are defined by the squared \emph{inverse distances}
\begin{subequations}
\begin{equation}
    w_k(x)=\frac{1}{d^2(x,x_k)}
\label{eq:w-IDW-basic}
\end{equation}
In order to make the weight decay more quickly as $x$ gets 
more distant from $x_k$, as suggested in~\cite{JK11,Bem20},
here we adopt the alternative weighting function
\begin{equation}
    w_k(x)=\frac{e^{-d^2(x,x_k)}}{d^2(x,x_k)}
\label{eq:w-IDW}
\end{equation}
\end{subequations}
Then, we define the following functions $v_k:\rr^n\to\rr$
for $k=1,\ldots,N$ as
\begin{equation}
    v_k(x)=\left\{\ba{ll}
    1&\mbox{if}\ x=x_k\\
    0&\mbox{if}\ x=x_j,\ j\neq k\\
    \displaystyle{\frac{w_k(x)}{\sum_{j=1}^Nw_k(x)}} & \mbox{otherwise}\ea\right.
\label{eq:v}
\end{equation}

As suggested in~\cite{JK11,Bem20}, we then define $s^2:\rr^n\to\rr^m$ as the 
\emph{IDW variance function}
\begin{equation}
    s_i^2(x)=\sum_{k\in\QQ} v_k(x)([y_k]_i-[\hat y(x)]_i)^2,\ i=1,\ldots,m
\label{eq:confidence}
\end{equation}
associated with the current training dataset $\{(x_k,y_k)\}_{k=1}^N$ and predictor $\hat y$.
Note that for $x=x_k$ and $k\in\QQ$ we have $s_i^2(x_k)=([y_k]_i-[\hat y(x_k)]_i)^2$, which in 
the case of perfect interpolation $[\hat y(x_k)]_i=[y_k]_i$ gives $s_i^2(x_k)=0$ (this corresponds to having no prediction uncertainty
about $y_i(x)$ at $x=x_k$). 
Note also that the sum in~\eqref{eq:confidence} only considers the indices
$k\in\QQ$, as for $k\not\in\QQ$ vector $x_k\not\in\XX$ and therefore
$y_k=y(x_k)$ is undefined. This is equivalent to assume that $y_k=\hat y(x_k)$
for all $x_k\not\in\XX$ and sum in~\eqref{eq:confidence} for $k=1,\ldots,N$.

Regarding promoting diversity in exploring the feature-vector space, as suggested in~\cite{Bem20} 
we also consider the \emph{IDW exploration function} $z:\rr^n\to\rr$ defined as
\begin{equation}
    z(x)=\left\{\ba{ll}
0 & \mbox{if}\ x\in\{x_1,\ldots,x_N\}\\
\frac{2}{\pi}\tan^{-1}\left(\frac{1}{\sum_{k=1}^Nw_k(x)}\right)&
\mbox{otherwise}\ea\right.
\label{eq:IDW-function}
\end{equation}
Similarly to the passive sampling approach in~\cite{YK10}, function $z$ returns a pure exploration term that is only based on the geometric position of
the (scaled) feature vectors $\{x_k\}$, and hence, contrarily to the IDW variance function $s^2$, 
is not influenced by the predictor $\hat y$ learned up to step $N$. Note that
$s^2$ also promotes exploration, but only indirectly.

\begin{figure}[t]
    \centerline{\includegraphics[width=\hsize]{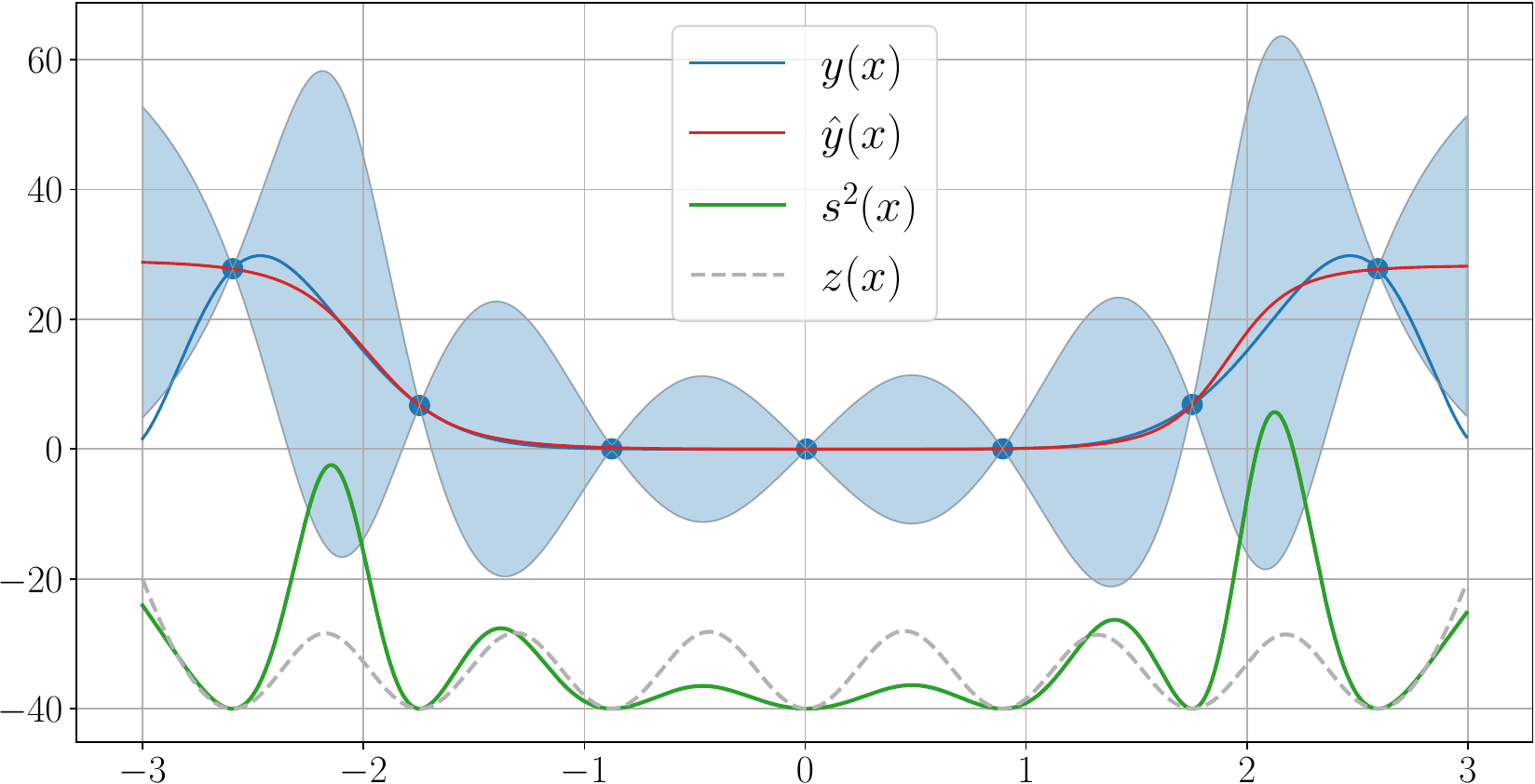}}
    \caption{Function $y$ of Example~\ref{ex:1d-ideal} (blue line), samples $(x_k,y_k)$ (blue dots), NN predictor
    $\hat y$ (red line), band $\hat y(x)\pm 3\sqrt{s_i^2(x)}$ (light blue area),
    scaled and shifted IDW functions $s^2$ (green line) and $z$ (dashed gray line)    
    }
    \label{fig:1d-ideal-1}
\end{figure}

\example{ex:1d-ideal}{
Let the data $y_k$ be generated by the following 
scalar function $y:\rr\to\rr$
\begin{equation}
    y(x)=x^4\sin^2\left(\frac{1}{3}x^2\right)
    \label{eq:fun-ideal}
\end{equation}
that we want to approximate over the interval $\bar\XX=[-3,3]$
by a simple feedforward neural network (NN) $\hat y$ with two layers of five neurons
each, logistic activation function
$\frac{1}{1+e^{-x}}$, and linear output function. 
As depicted in Figure~\ref{fig:1d-ideal-1},
we assume that we have collected $N=7$ samples $(x_k,y_k)$
(blue dots), $y_k=y(x_k)$, and fit a NN via the 
\texttt{MLPRegressor} function in \texttt{scikit-learn}~\cite{scikit-learn}
with $\ell_2$-regularization term $\alpha=10^{-2}$, by using the
L-BFGS nonlinear optimization algorithm~\cite{LN89}.
Figure~\ref{fig:1d-ideal-1} also shows 
the original function $y(x)$ (blue line), the NN predictor
$\hat y(x)$ (red line), and the band $\hat y(x)\pm 3\sqrt{s_i^2(x)}$ (light blue area).
The figure also shows scaled and shifted versions of the IDW functions $s^2(x)$ 
(green line) and $z(x)$ defined in~\eqref{eq:IDW-function} (dashed gray line).
}

Let us now define the acquisition function
\begin{equation}
    a(x)=\left(\sum_{i=1}^m s_i^2(x)\right)+\delta z(x)
\label{eq:acquisition_fcn_base}
\end{equation}
where $\delta\geq 0$ is a hyperparameter balancing the role of IDW variance $s^2(x)$
and IDW distance $z(x)$. Note that $\delta$
trades off between model-based learning (small $\delta$) and 
learning based on the pure exploration of the feature-vector space 
to promote diversity (large $\delta$).

In the case of population-based AL, the maximization problem~\eqref{eq:acquisition}
can be solved by global optimization; in this paper, we will use
the derivative-free Particle Swarm Optimization (PSO) algorithm~\cite{KE95},
as $a(x)$ is a cheap function to evaluate
whenever $\hat y(x)$ is easy to compute. 
In pool-based sampling, when
the number $M$ of samples in the pool is not too high,
problem~\eqref{eq:acquisition} can be solved by enumeration by setting
\begin{subequations}
\begin{equation}
    x_{N+1}=\bar x_{k^*},\ k^*=\arg\min_{k\in\{1,\ldots,M\}\setminus\EE}\{a(\bar x_k)\}    
\label{eq:xN+1-a}
\end{equation}
We assume that possible duplicates $\bar x_k=\bar x_j$, $k\neq j$, are removed upfront from 
the pool $\XX_P$.

When~\eqref{eq:xN+1-a} is impractical due to a large number $M$ of samples in the pool,
one can first use PSO to optimize over the entire set $\bar\XX$ to get $\bar x^*=\arg\max_{x\in\bar\XX}a(x)$ as in population-based AL 
and then set (cf.~\cite{WHYL15})
\begin{equation}
    x_{N+1}=\bar x_{k^*},\ k^*=\arg\min_{k\in\{1,\ldots,M\}\setminus\EE} d^2(\bar x_k,x^*)
\label{eq:xN+1-b}
\end{equation}
\label{eq:xN+1}%
\end{subequations}

Algorithm~\ref{algo:IDEAL} reports the pseudocode of the proposed AL algorithm
that we call Inverse-Distance based Exploration for Active Learning (IDEAL). The complexity of the algorithm will be discussed in Section~\ref{sec:complexity}. Note that output data scaling can be updated before retraining $\hat y$ at Step~\ref{algo:ideal:retrain}, such as by applying standard scaling based on the currently available values $\{y_k\}$, $k\in\QQ$. 

\begin{algorithm}[h!]
    \caption{Inverse-Distance based Exploration for Active Learning (IDEAL).}
    \label{algo:IDEAL}
    ~~\textbf{Input}: Set $\XX_P=\{\bar x_k\}_{k=1}^M$ (pool-based) or $\XX_P=\bar\XX$ (population-based) of queryable feature vectors; 
budget $N_{\rm max}$ of available queries; number $N_i$ of initial
samples to acquire; pure exploration hyperparameter $\delta\geq 0$.
    \vspace*{.1cm}\hrule\vspace*{.1cm}
    \begin{enumerate}[label*=\arabic*., ref=\theenumi{}]
        \item Remove possible duplicates $\bar x_k$ from $\XX_P$ (pool-based only);
        \item Compute scaling functions $\sigma_i$ as in~\eqref{eq:scaling};
        \item Extract $N_i$ samples $(x_k,y_k)$ as described
        in Section~\ref{sec:init} by K-means (pool-based) or LHS (population-based);
        if it is not possible to extract them within $N_{\rm max}$ queries 
        go to Step~\ref{algo:ideal:stop}, otherwise set $N_{\rm init}$ = number of queries done;
        \label{algo:ideal:init}
        \item $\QQ\leftarrow\{k\in\{1,\ldots,N_{\rm init}\}:\ x_k\in\XX\}$;
        \item $\EE\leftarrow\big\{i\in\{1,\ldots,M\}:\ \bar x_i=x_k$ for some $k\in\{1,\ldots,N_{\rm init}\}\big\}$ (pool-based only);
        \item \textbf{For} $N=N_{\rm init},\ldots,N_{\rm max}$ \textbf{do}:
        \begin{enumerate}[label=\theenumi{}.\arabic*., ref=\theenumi{}.\arabic*]
                \item \textbf{If} $N\not\in\QQ$ \textbf{then} update predictor $\hat y$ by solving~\eqref{eq:supervised_learning-emp};\label{algo:ideal:retrain}
                \item Compute new sample $x_{N+1}$ as in~\eqref{eq:acquisition}
                (population-based) or~\eqref{eq:xN+1} (pool-based);\label{algo:ideal:acquisition}
                \item \textbf{If} $x_{N+1}\in\XX$ acquire $y_{N+1}$ and set
                $\QQ\leftarrow\QQ\cup\{N+1\}$;
                \item $\EE\leftarrow\EE\cup\{k^*\}$ (pool-based only);
        \end{enumerate}
        \item \textbf{End}.\label{algo:ideal:stop}
    \end{enumerate}
    \vspace*{.1cm}\hrule\vspace*{.1cm}
    ~~\textbf{Output}: Predictor $\hat y$, or declaration of failure in collecting $N_i$ feasible initial samples.
\end{algorithm}

\subsection{Extensions of the acquisition function}
The basic acquisition function~\eqref{eq:acquisition_fcn_base} can be extended in two
directions. First, for pool-based AL we can consider the density function $\rho:\XX_P\to(0,+\infty)$
that measures how much ``isolated'' is a sample $\bar x_k\in\XX_P$ with respect
to the remaining samples. Similar to density-based spatial clustering approaches~\cite{EKSX96},
one can use the average (scaled) distance of $\bar x_k$ from its $n$ nearest neighbors, 
\[
    d_k=\frac{1}{n}\sum_{j\in \NN_k} \|\sigma(\bar x_k)-\sigma(\bar x_j)\|_2
\]
where $\NN_k$ is the set of indices corresponding the $n$ nearest neighbors of $\bar x_k$ in $\XX_P\setminus\{\bar x_k\}$, to estimate the density as proportional to the normalized inverse
volume of the sphere of radius $d_k$, i.e.,
\begin{equation}
    \rho(\bar x_k)=\frac{\frac{1}{d_k^n}}{\max_{j=1,\ldots,M}\left\{\frac{1}{d_j^n}\right\}}=
    \frac{\min_{j=1,\ldots,M}\left\{d_k^n\right\}}{d_k^n}    
\label{eq:rho_k}
\end{equation}
Note that~\eqref{eq:rho_k} is always defined, as $n$ duplicates cannot exist
such that they have a zero average distance due to the fact that we have assumed that all possible duplicates $\bar x_k=\bar x_j$, $k\neq j$, have been removed.
Note that $\rho$ does not depend on the predictor $\hat y$ learned and can be therefore
computed upfront. Regarding population-based AL, one can simply set $\rho(x)=1$, $\forall x\in\bar\XX$.

Next, we can introduce weight functions $c_i:\rr^n\to[0,+\infty)$  to actively learn the predictor in a non-uniform way with respect to the target index $i$ and $x$
(or uniformly, if $c_i(x)\equiv 1$ for all $i=1,\ldots,m$). Accordingly, we extend~\eqref{eq:acquisition_fcn_base} to
\begin{equation}
    a(x)=(1+\omega\rho(x))\sum_{i=1}^mc_i(x)\left(s_i^2(x)+\frac{\delta}{m} z(x)\right)
\label{eq:acquisition_fcn_full}
\end{equation}
where $\omega\geq 0$ is a scalar weight on density. Note that $\omega$ is redundant
in the case of population-based AL, having assumed that $\rho(x)\equiv 1$.

Let us show that the active learning mechanism~\eqref{eq:acquisition} under~\eqref{eq:acquisition_fcn_base}, possibly extended as in~\eqref{eq:acquisition_fcn_full}, follows criteria of \emph{informativeness}, \emph{representativeness},
and \emph{diversity}, which are listed in~\cite{Wu19} as essential for AL. 
Regarding the first, maximizing $a(x)$ implies looking for large values of the uncertainty $s(x)$ associated with the current predictor $\hat y(x)$, i.e., to select 
the next sample $x_{N+1}$
where $\hat y$ is considered most uncertain according to~\eqref{eq:confidence}, so that querying $x_{N+1}$ is 
expected to bring significant new information. The second, which is only applicable
in the case of pool-based sampling under the extension~\eqref{eq:acquisition_fcn_full}, is taken care of by $\rho(x)$ when $\omega>0$, as in the maximization~\eqref{eq:acquisition} those samples $\bar x_k$ that have a low density $\rho(\bar x_k)$, for instance, because they are outliers, will be discouraged.
Third, diversity is promoted because $s(x)$ and $z(x)$ are small close to samples
that have been already visited, which ultimately makes the AL algorithm visit
unexplored areas of the feature-vector space. The tradeoff between representativeness and diversity is taken care of by the coefficient $\omega$.

In all the numerical tests reported in Section~\ref{sec:results} we will always employ the 
baseline acquisition function~\eqref{eq:acquisition_fcn_base}, as no significant improvements were found
by using $\omega>0$ in our benchmarks, and in addition we aim at a uniform weighting $c_i(x)\equiv 1$. Nonetheless, the extra versatility allowed by~\eqref{eq:acquisition_fcn_full} might be useful in certain AL applications.

\subsection{Other active learning algorithms}
\label{sec:other-AL}
Algorithm~\ref{algo:IDEAL} (\ideal) will be compared to some of the most common active learning methods proposed in the literature that can support rather arbitrary prediction models $\hat y$.
The considered methods have some substantial differences, that we will see
have consequences on AL performance. We review such methods here below.

\subsubsection{Random sampling (\random)}
The method draws  samples $x_{N+1}$ from the uniform distribution defined over $\bar\XX$ in the case of population-based sampling, or by selecting a random index in $\{1,\ldots,N_{\rm max}\}\setminus\EE$ in the case of pool-based sampling. This is the simplest method we consider to have a baseline to compare with: any AL method should be more efficient that {\random}, at least statistically.

\subsubsection{Greedy method (\greedyx)}
The sampling technique {\greedyx} proposed in~\cite[Algorithm 1]{YK10} selects $x_{N+1}$ by maximizing the minimum distance from existing samples, i.e., 
\begin{subequations}
\beqar
    x_{N+1}&=&\arg\max_{x\in\XX^P}d_x(x)\label{eq:greedy_x-min}\\
    d_x(x)&=&\min_{k=1,\ldots,N}\|\sigma(x)-\sigma(x_k)\|_2^2\label{eq:d_x}
\eeqar%
\label{eq:greedy_x}%
\end{subequations}
The method is not model-based, in that the predictor $\hat y$ is not used
to select the samples to query.
Although conceived for pool-based AL, the method can be extended also
to population-based AL by maximizing $d(x)$ with respect to $x\in\bar\XX$ 
in~\eqref{eq:greedy_x}. For fair comparison, in the case of population-based AL,
rather than maximizing the minimum distance in our numerical tests 
we will also adopt LHS to acquire the first $N_i$ samples instead
of using the approach suggested in~\cite{WLH19} for pool-based AL.

\subsubsection{Greedy method (\greedyxy)}
The {\greedyxy} method proposed in~\cite[Algorithm 3]{WLH19} is an extension of greedy sampling 
that, in addition, considers the minimum predicted distance in the $y$-space 
\begin{equation}
    d_y(x)=\min_{k=1,\ldots,N}\|\hat y(x)-y_k\|_2^2
\label{eq:d_y}
\end{equation}
where $\hat y$ is the latest predictor trained on currently available samples,
and selects
\begin{equation}
    x_{N+1}=\arg\max_{x\in\XX^P}d_x(x)d_y(x)
\label{eq:greedy_xy}
\end{equation}
(in case of multiple targets, we assume that the values in~\eqref{eq:d_y} refer to scaled target values). The method can be extended also to population-based AL. In such a case, similarly to {\greedyx}, 
we will use LHS for initialization.

Thanks to the additional term $d_y$ defined in~\eqref{eq:d_y}, {\greedyxy} also
aims at getting samples where the output $y$ is expected to be different from
the current values $y_k$ observed so far. 
A possible drawback of~\eqref{eq:greedy_xy}, however, is that $d_x$ and $d_y$ are multiplied
by each other, i.e., diversity is sought in both the $x$- \emph{and} $y$-space, so that 
pure exploration of the $x$-space
might be inhibited by \emph{predicted} proximity in the $y$-space, i.e., by 
small (or zero) estimated values $d_y$. Instead, {\ideal} looks for diversity in
the $x$- \emph{or} $y$-space, as $z(x)$ and $s^2(x)$ are summed in~\eqref{eq:acquisition_fcn_base} instead of being multiplied by each other as in~\eqref{eq:greedy_xy}.

\subsubsection{Query-by-Committee (\QBC)}
After a first initialization phase in which $N_i$ feasible samples
are generated randomly, the QBC method for regression~\cite{RH95,BRK07} considered here
creates $K_{\textrm QBC}$ bootstrap samples obtained by randomly sampling
the existing $N$ samples with replacement, trains a predictor $\hat y^j$ on each set, $j=1,\ldots,K_{\textrm QBC}$, and then selects $x_{N+1}$ to maximize the output-prediction variance
\begin{equation}
    x_{N+1}=\arg\max_{x\in\XX^P}\sum_{j=1}^{K_{\textrm QBC}}\left\|\hat y^j(x)-
    \frac{1}{K_{\textrm QBC}}\sum_{j=1}^{K_{\textrm QBC}}\hat y^j(x)
    \right\|_2^2
\label{eq:QBC}
\end{equation}
(in case of multiple targets, the terms in~\eqref{eq:QBC} must be considered again
as scaled target values).
It can be used for both pool-based and population-based AL.
In {\QBC}, we set $K_{\textrm QBC}=5$ in all our tests and train the individual predictors $\hat y^j$ on bootstrapped samples as in~\cite{WLH19},
rather than leaving out a different subset of $\lfloor \frac{N}{K_{\textrm QBC}}\rfloor$ samples as suggested in~\cite{BRK07}. In fact, the former approach better performed in our examples, in which the number of allowed samples is small compared to the number of model parameters to learn and hence removing $\lfloor \frac{N}{K_{\textrm QBC}}\rfloor$ samples can dramatically change the resulting individual 
predictions $\hat y_j(x)$. An additional disadvantage of {\QBC} when actively learning NN models is that large disagreements may be caused by lack of global convergence of the optimization method used to train the different predictors, due to the non-convex nature of the training problem.

The {\QBC} method we tested only relies on information related to the $y$-space, i.e.,
is totally based on the predictor $\hat y$ and its variants $\hat y^j$
to drive the acquisition, but not explicitly on measures defined purely on
feature-vectors for promoting diversity in the $x$-space.
This ultimately has a potential negative impact on the robustness of {\QBC}.
Extensions of {\QBC} to improve performance by taking into account diversity and density was introduced in~\cite{KDR18} in the context of classification.

\subsubsection{Improved Representativeness-Diversity Maximization ({\iRDM})}
The {\iRDM} pool-based unsupervised active learning method~\cite{LJLFLW21} for regression generates $N_{\rm max}$ samples in one shot
(rather than incrementally) after performing K-means~\cite{Llo57} on $\XX^\sigma_P$
to create $N_{\rm max}$ clusters. Then, the samples closest to the resulting centroids are
refined sequentially (up to $c_{\rm max}$ times) to optimize the tradeoff between the representativeness of the selected point $x_k$ within its cluster $C_k$ (i.e., the average distance of $x_k$ from the points in $C_k$) and the diversity of $x_k$ from the other selected samples $x_1,\ldots,x_{k-1},x_{k+1},\ldots,x_{N_{\rm max}}$ in the remaining clusters (i.e., the minimum distance $\|\sigma(x_k)-\sigma(x_j)\|_2$, $j=1,\ldots,N_{\rm max}$, $j\neq k$). We set $c_{\textrm max}=5$ in all our tests as suggested in~\cite{LJLFLW21}. As for {\greedyx}, the method does not exploit the predictor $\hat y$, which in fact is only trained after acquiring all the $N_{\rm max}$ samples.

\subsection{Numerical complexity}
\label{sec:complexity}
The initial phase of {\ideal} (Algorithm~\ref{algo:IDEAL}) requires either extracting $N_{\rm init}$
samples by Latin Hypercube Sampling (LHS)~\cite{MBC79} (population-based AL)
or K-means~\cite{Llo57} (pool-based AL). In addition, {\ideal} 
requires retraining the predictor $\hat y$ at Step~\ref{algo:ideal:retrain} 
and solving the optimization problem at Step~\ref{algo:ideal:acquisition}
to get a new sample. Depending on the class of predictors $\hat y$ used, retraining can be the most expensive computation effort. As for other AL methods, warm-starting the training algorithm
or incrementally learning $\hat y$ could be exploited, if supported by the particular
class of prediction models and training algorithms chosen. Regarding Step~\ref{algo:ideal:acquisition}, the computation complexity mainly depends on the number of operations required to evaluate the predictor $\hat y(x)$ in~\eqref{eq:confidence}, and therefore on the complexity of the selected model class.

Algorithms {\ideal}, {\greedyxy}, and {\QBC} require retraining the predictor $\hat y$,
respectively, $N_{\rm max}-N_{\rm init}+1$, $N_{\rm max}-N_{\rm init}+1$, and $(K_{\textrm QBC}+1)(N_{\rm max}-N_{\rm init}+1)$ times, while {\random}, {\greedyx}, and {\iRDM} only once at the
end of the acquisition, as they are only based on the relative positions of
the acquired samples $x_k$ in the feature space.

In the case of pool-based sampling, {\ideal}, {\greedyxy},
and {\QBC} require, after the initialization phase, evaluating the predictor $\hat y$, respectively, $(N_{\rm max}-N_{\rm init}+1)M$, $(N_{\rm max}-N_{\rm init}+1)M$, and $(K_{\textrm QBC}+1)(N_{\rm max}-N_{\rm init}+1)M$ times. In addition, {\ideal} requires
evaluating the acquisition function $a(\bar x_k)$ $(N_{\rm max}-N_{\rm init}+1)M$ times, {\greedyx} and
{\greedyxy} evaluating the squared distances in~\eqref{eq:d_x}
$(N_{\rm max}-N_{\rm init}+1)M$ times, and in addition {\greedyxy} evaluating 
the squared distances~\eqref{eq:d_y}
$(N_{\rm max}-N_{\rm init}+1)M$ times, while {\QBC} requires evaluating
the output prediction variance terms in~\eqref{eq:QBC}
$(N_{\rm max}-N_{\rm init}+1)M$ times. Then, {\ideal}, {\greedyx}, {\greedyxy}, and 
{\QBC} require computing $N_{\rm max}-N_{\rm init}+1$ times the minima
defined by~\eqref{eq:xN+1-a},~\eqref{eq:greedy_x-min},~\eqref{eq:greedy_xy},
and~\eqref{eq:QBC}, respectively. 

Regarding {\iRDM}, it requires solving 
K-means to partition $M$ points into $N_{\rm max}$ clusters,
computing the representativeness measure $M$ times, and then 
repeat $c_{\rm max}$ times the construction of the diversity measure
on all candidate samples within the updated cluster with respect to the  
samples already fixed in each one of the other clusters. Note that {\iRDM}, contrarily to the other
methods, is not an incremental AL method, and 
therefore a change of $N_{\rm max}$ (such as due to allowing more queries) 
would require redefining all the $N_{\rm max}$ samples to acquire.

Finally, we remark that the computation time required by the AL algorithm 
is often negligible with respect to the time required to acquire a new target value,
which is often the most dominating effort in the practical situations
AL algorithms are employed.

\section{Numerical tests}
\label{sec:results}
In this section, we test the proposed AL approach on synthetic illustrative examples
and real-world datasets, comparing it to the different AL methods reviewed
in Section~\ref{sec:other-AL}. 

Regarding the initial $N_i$ samples, for the {\greedyx} and {\greedyxy}
methods we recursively use~\eqref{eq:d_x} starting from the centroid $x_1$ of $\XX^P$
as proposed in~\cite{WLH19}, and random sampling for the {\random} and {\QBC} methods.
Different approaches have been proposed in the literature for cold-starting AL,
see for instance the representative sampling method proposed recently in~\cite{JYLWWS22} in the context of image classification, or the approach used by {\iRDM}. 

To analyze the performance of {\iRDM} as a function of the number of acquired samples, 
as {\iRDM} is not an incremental method we execute it from scratch 
each time we want to collect a different number of samples.

All computations were carried out
in Python 3.9.15 using the \textsf{scikit-learn} package~\cite{scikit-learn} to
train feedforward NNs for regression (\texttt{MLPRegressor} function)
and support vector regression (SVR) (\texttt{SVR} function).
Regarding the considered AL methods, for pool-based active learning we used 
the Python implementation developed by the author and available at {\texttt{\url{http://cse.lab.imtlucca.it/~bemporad/ideal}}}.

\begin{figure}[t]
    \centerline{\includegraphics[width=.8\hsize]{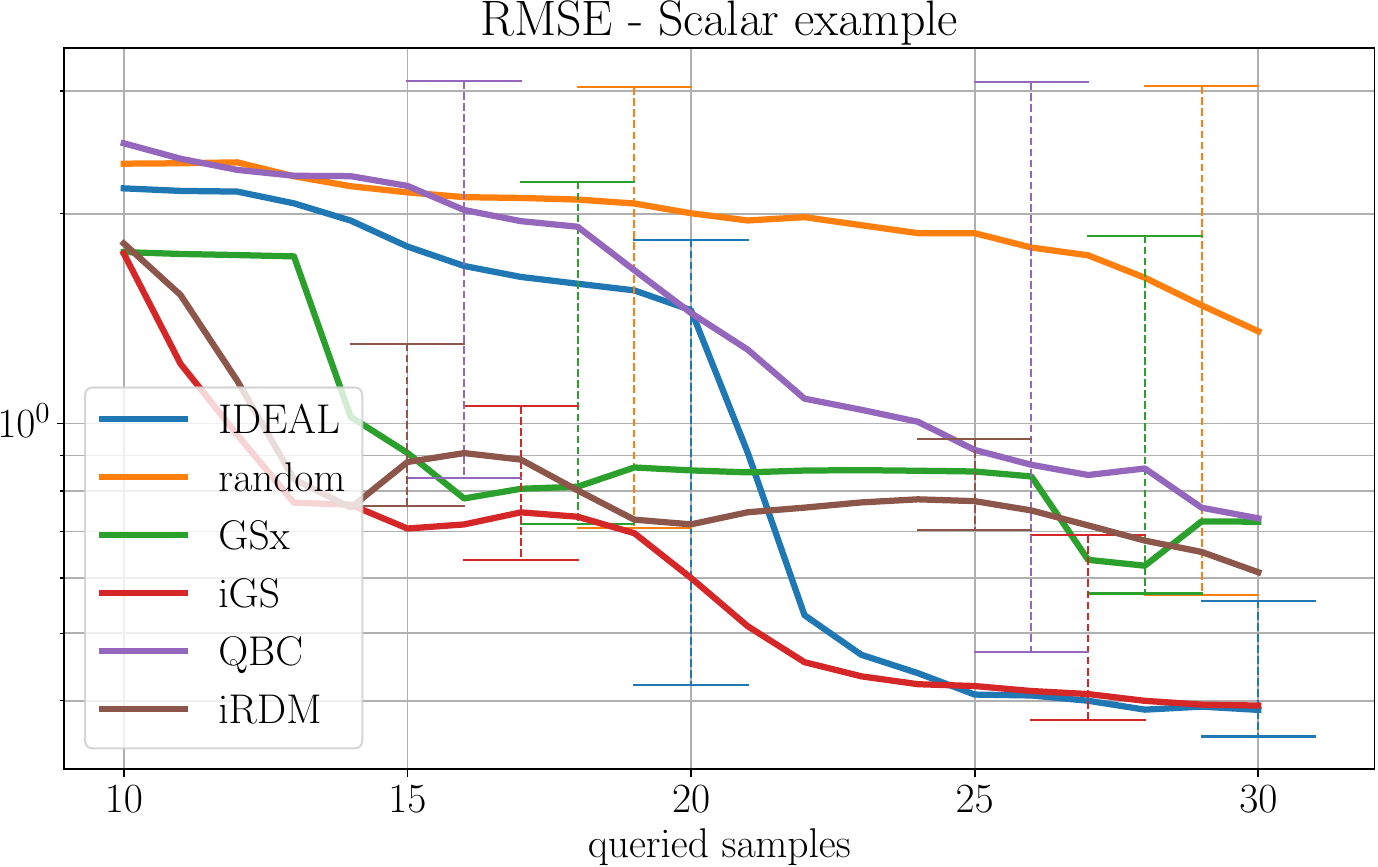}}
    \caption{AL of function~\eqref{eq:fun-ideal}: median RMSE as a function of the number of queries. Vertical lines denote min and max RMSE values}
    \label{fig:1d-ideal-2}
\end{figure}

\subsection{Scalar example}
We first test the proposed AL approach on the simple regression problem defined in Example~\ref{ex:1d-ideal}, i.e., with $y(x)$ as in~\eqref{eq:fun-ideal}. Since $n=1$, we generate a grid $\XX_P$ of $M=1000$ equally-spaced points on the line segment $\bar\XX=[-3,3]$ and use pool-based sampling on the entire pool $\XX_P$, so that problem~\eqref{eq:acquisition} can be solved by enumeration~\eqref{eq:xN+1-a}. 
While training the NN, the parameter vector is not warm-started when executing Step~\ref{algo:ideal:retrain}, to avoid possible low-quality local minima inherited by the early steps of Algorithm~\ref{algo:IDEAL} when only a few data are available.

The median over 50 runs of the root-mean-square error (RMSE) 
\[
    RMSE=\sqrt{\frac{1}{M}\sum_{k=1}^{M}(y_k-\hat y(x_k))^2}
\]
and its range (min and max values) 
obtained with $\delta=5$, $N_i=10$, $N_{\rm max}=30$, as a function of the number
$N$ of acquired samples, is depicted in Figure~\ref{fig:1d-ideal-2} and compared
with the RMSE obtained with \random, \greedyx,
\greedyxy, {\QBC}, and {\iRDM} sampling. It is apparent that {\ideal} is superior
to {\random} and {\QBC}, behaves better than {\greedyx} and {\iRDM} (which
are not model-based methods) and similarly to {\greedyxy}
after about half of the allowed samples have been acquired. The high variance of {\QBC} is possibly due to the small number of samples queried and, consequently, the even smaller number
of training samples used to train the predictors forming the committee that may lead
to large disagreements among them.

Table~\ref{tab:1d-ideal} shows the mean and standard deviation of the RMSE obtained 
by {\ideal} when $N=N_{\rm max}$ for different values of the hyperparameter $\delta$
and, for comparison, by {\random} sampling. 

\begin{table}[h]
    \begin{center}{\small
    \begin{tabular}{r|rrrrr|r}
    $\delta$& 0.0& 0.1& 1.0& 5.0& 10.0& R  \\\hline mean &
    0.528 &0.470 &0.439 &0.402 &0.402 &1.495 \\\hline
    std &
    0.307 &0.239 &0.084 &0.042 &0.036 &0.548 \\\hline
    \end{tabular}}
\end{center}
    \caption{AL of function~\eqref{eq:fun-ideal}: mean RMSE and its standard deviation after $N_{\rm max}=30$ steps obtained on 50 different runs of Algorithm~\ref{algo:IDEAL} for different values of $\delta$ (R = {\random} sampling)} 
    \label{tab:1d-ideal}
\end{table}

We will take $\delta=5$ in all our remaining tests.
For such a value of $\delta$, to test the robustness of AL against measurement noise we repeat
the same test by perturbing the measurements $y_k=y(x_k)+\eta_k$, where $\eta_k\sim\NN(0,\sigma_\eta^2)$ for different values of the standard deviation $\sigma_\eta$.
The mean and standard deviation of the resulting RMSE over 50 runs after $N_{\rm max}=30$ 
iterations is shown in Table~\ref{tab:RMSE-ideal-noise}. The table shows that 
for increasing values of $\sigma_\eta$ the RMSE deteriorates without an excessive
increase of variance and in a gradual way for {\ideal} (proving its robustness with respect to noisy target measurements), \greedyxy, and {\iRDM}, while such a trend is less marked for {\greedyx}, {\random}, and {\QBC}.

\begin{table}[h]
    \begin{center}{\small
    \begin{tabular}{c|c|c|c|c|c|c}
    $\sigma_\eta$& \ideal & \random & \greedyx & \greedyxy & \QBC & \iRDM\\\hline
    $0.0$ &  0.40 (0.04) &  1.44 (0.60) &  0.76 (0.14) &  0.41 (0.04) &  1.16 (0.88) &  0.63 (0.12)\\
    $1.0$ &  0.62 (0.02) &  1.45 (0.55) &  0.80 (0.11) &  0.62 (0.04) &  1.15 (0.79) &  0.74 (0.16)\\
    $2.0$ &  0.86 (0.09) &  1.60 (0.48) &  0.92 (0.14) &  0.86 (0.05) &  1.38 (0.78) &  0.91 (0.14)\\
    \hline
    \end{tabular}}
    \end{center}
    \caption{AL of function~\eqref{eq:fun-ideal} with noise: mean (std) RMSE after $N_{\rm max}$ =30
    steps for different values of $\sigma_\eta$ 
    }
    \label{tab:RMSE-ideal-noise}
\end{table}

\subsection{Multiparametric quadratic programming}
Model predictive control (MPC) is a popular engineering
technique for controlling dynamical systems in an optimal way under
operating constraints~\cite{BBM17}. Evaluating the MPC law requires solving a
quadratic programming (QP) problem of the form
\begin{equation}
    \begin{array}{lcll}
        z^*(x)&=&\arg\min_z & 
                 \hspace*{-1em}\frac{1}{2}z'Qz+x'F'z\\
        && \st & \hspace*{-1em}Az\leq b+Sx\\
        &&    &\hspace*{-1em} \ell\leq z\leq u\\[.5em]
        y(x)&=&[I_m\ 0\ \ldots\ 0]&
                    z^*(x)
    \end{array}
\label{eq:mpQP}
\end{equation}
where $z\in\rr^{n_z}$ is a vector of future control moves, $n_z\geq m$,
and $x\in\rr^n$ is a vector of parameters that change at run time,
such as estimated states and reference signals, and the Hessian matrix $Q=Q'\succ 0$. To alleviate the effort
of solving~\eqref{eq:mpQP} online for each given vector $x$, multiparametric QP (mpQP)
was proposed in~\cite{BMDP02a}, showing that the solution 
$z^*:\rr^n\to\rr^{n_z}$, and therefore $y(x)$, is continuous and piecewise affine over a polyhedral partition of a convex polyhedron $\XX\subseteq\rr^n$. The main drawback
of such an \emph{explicit} form of MPC is that the number of polyhedral cells
tends to grow exponentially with the number of constraints in~\eqref{eq:mpQP}.

Suboptimal methods were proposed to \emph{approximate} $y(x)$, such as via neural
networks~\cite{PZ95,KL20}. In order to find an approximation $\hat y(x)$ of $y(x)$, one must collect
a training dataset of pairs $(x_k,y_k)$, where evaluating $y_k=y(x_k)$ requires
solving a QP problem as in~\eqref{eq:mpQP}. Randomly sampling a given set $\bar\XX\subset\rr^n$ of parameters $x$ may result time-consuming, especially
when the dimension $n$ of the parameter vector is large. To minimize the number $N_{\rm max}$ of QP problems solved to get a proper approximation quality, we use Algorithm~\ref{algo:IDEAL} to actively generate samples $x_k$.

\begin{figure}[t]
    \centerline{\includegraphics[width=.8\hsize]{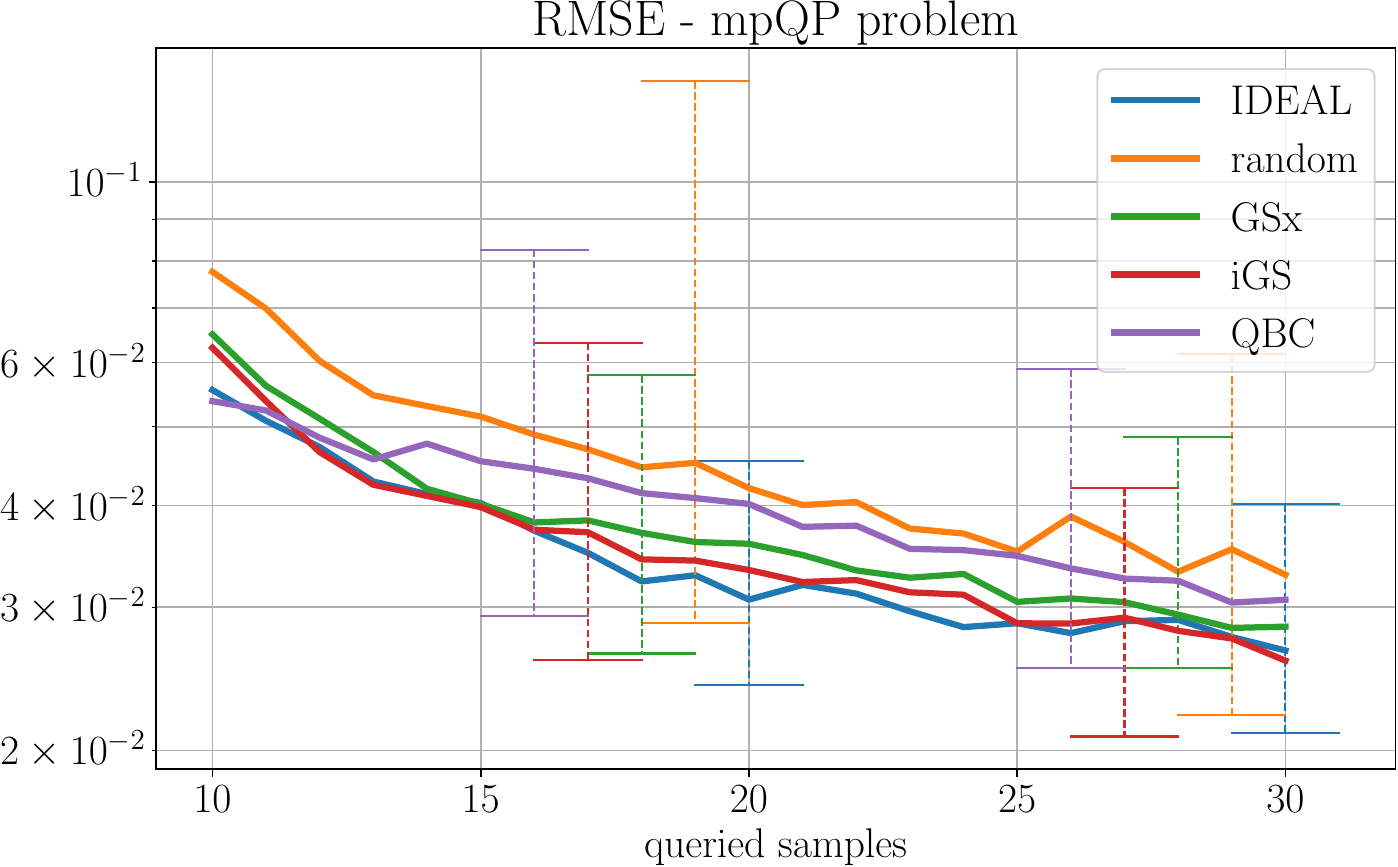}}
    \caption{mpQP problem: median RMSE as a function of the number of queries.
    Vertical lines denote min and max values. Only the population-based versions
of \ideal, \greedyx, \greedyxy, and {\QBC} were used ({\iRDM} is a pure pool-based method).}
    \label{fig:mpqp1-RMSE}
\end{figure}

\begin{figure}[t]
    \begin{center}
    \includegraphics[width=0.7\hsize]{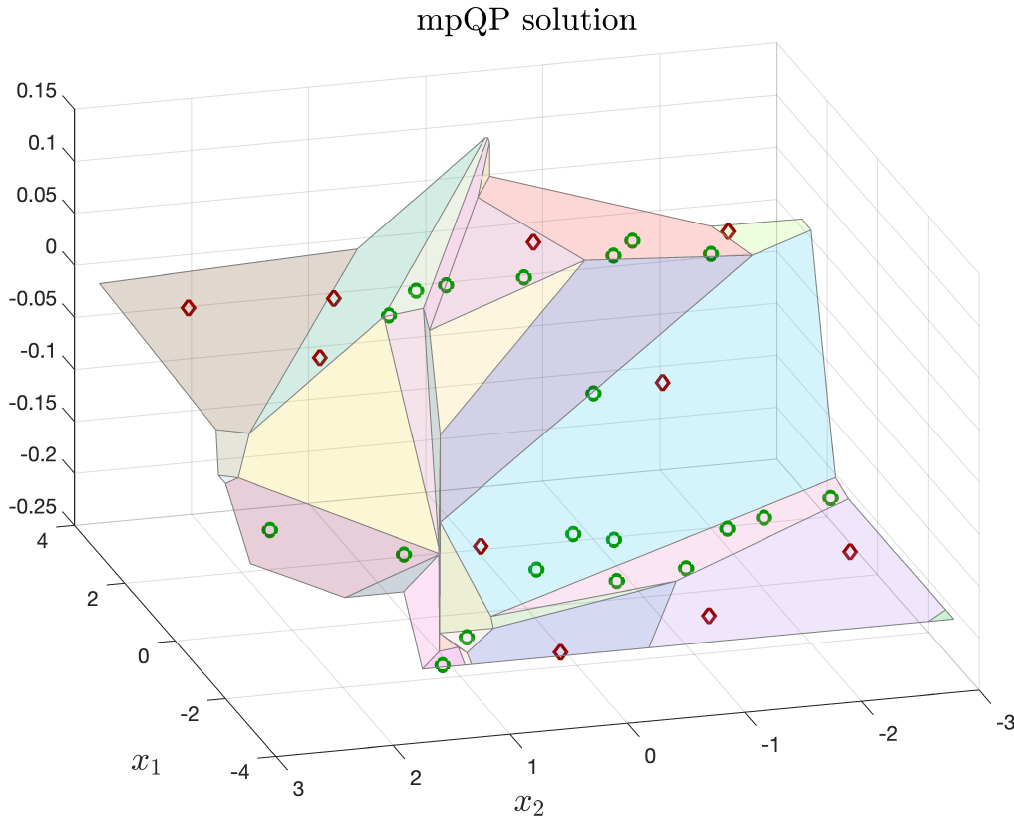}
    \end{center}
    \caption{Exact mpQP solution,
    initial samples $x_1,\ldots,x_{N_i}$
    (red diamonds), and samples 
$x_{N_i+1},\ldots,x_{N_{\rm max}}$ queried by {\ideal} (green circles)}
    \label{fig:mpqp1-partition}
\end{figure}

We consider here a mpQP problem with $n=2$, $n_z=12$, $m=1$, $b\in\rr^{12}$, $S=0$,
and all matrices in~\eqref{eq:mpQP} generated randomly, with the entries
of $A$, $F$ $\sim\NN(0,1)$ and the entries of $b,u,-\ell$ $\sim\UU[0,1]$,
where $\UU[0,1]$ is the uniform distribution over the interval $[0,1]$,
$Q=Q'\succ 0$, is randomly generated so that its condition number equals $10^3$,
and $\bar\XX=\{x: \|x_i\|_\infty\leq 3\}$. Algorithm~\ref{algo:IDEAL} is applied
using population-based sampling with $N_i=10$ and $N_{\rm max}=30$
for training a feedforward neural network with 3 layers of 10 neurons each and ReLU activation
function, without using warm starting while retraining the model. The median RMSE and its range over 50 runs is shown in Figure~\ref{fig:mpqp1-RMSE}, where it is apparent that {\ideal} performs better than \random, \greedyx, and {\QBC}, and similar to \greedyxy. 
Note that in this population-based AL example we could not use {\iRDM}, which is a pure pool-based method. 
Figure~\ref{fig:mpqp1-partition}
shows the polyhedral partition associated with the exact mpQP solution (unknown to the active learning algorithms) computed as described in~\cite{Bem15} along with the queried samples and initial samples generated by one of
the runs of Algorithm~\ref{algo:IDEAL}. It is evident that the points 
acquired by {\ideal} are not distributed uniformly.


\begin{figure}[th!]
    \begin{center}
    \includegraphics[width=0.5\hsize]{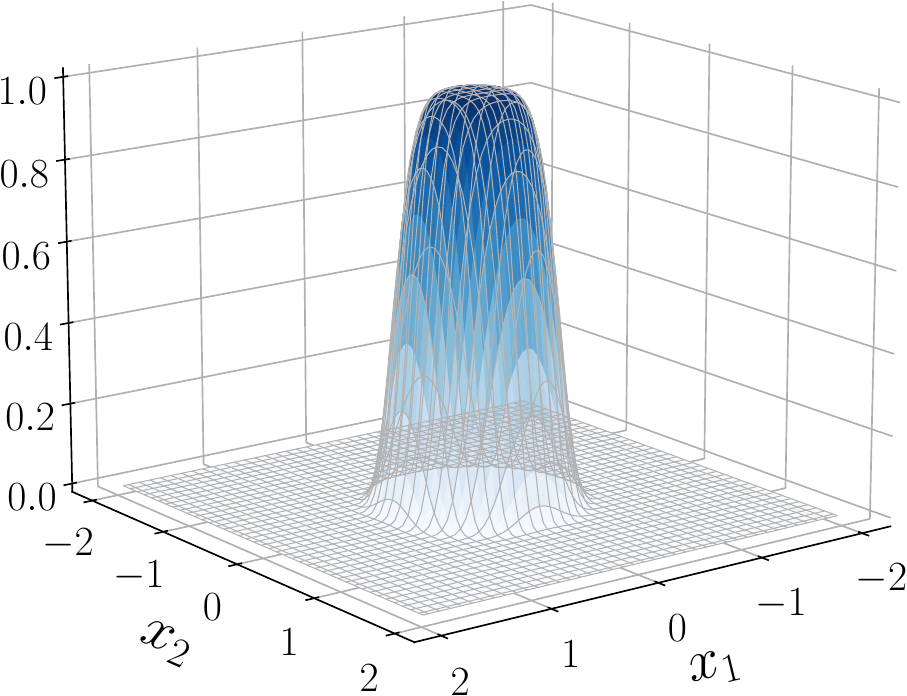}
    \end{center}
    \caption{Bell-shaped function~\eqref{eq:bell_fcn}}
    \label{fig:bell_fcn}
\end{figure}

\subsection{Active learning with unknown constraints}
\begin{figure*}[h!]
    \begin{center}
    \includegraphics[width=0.8\hsize]{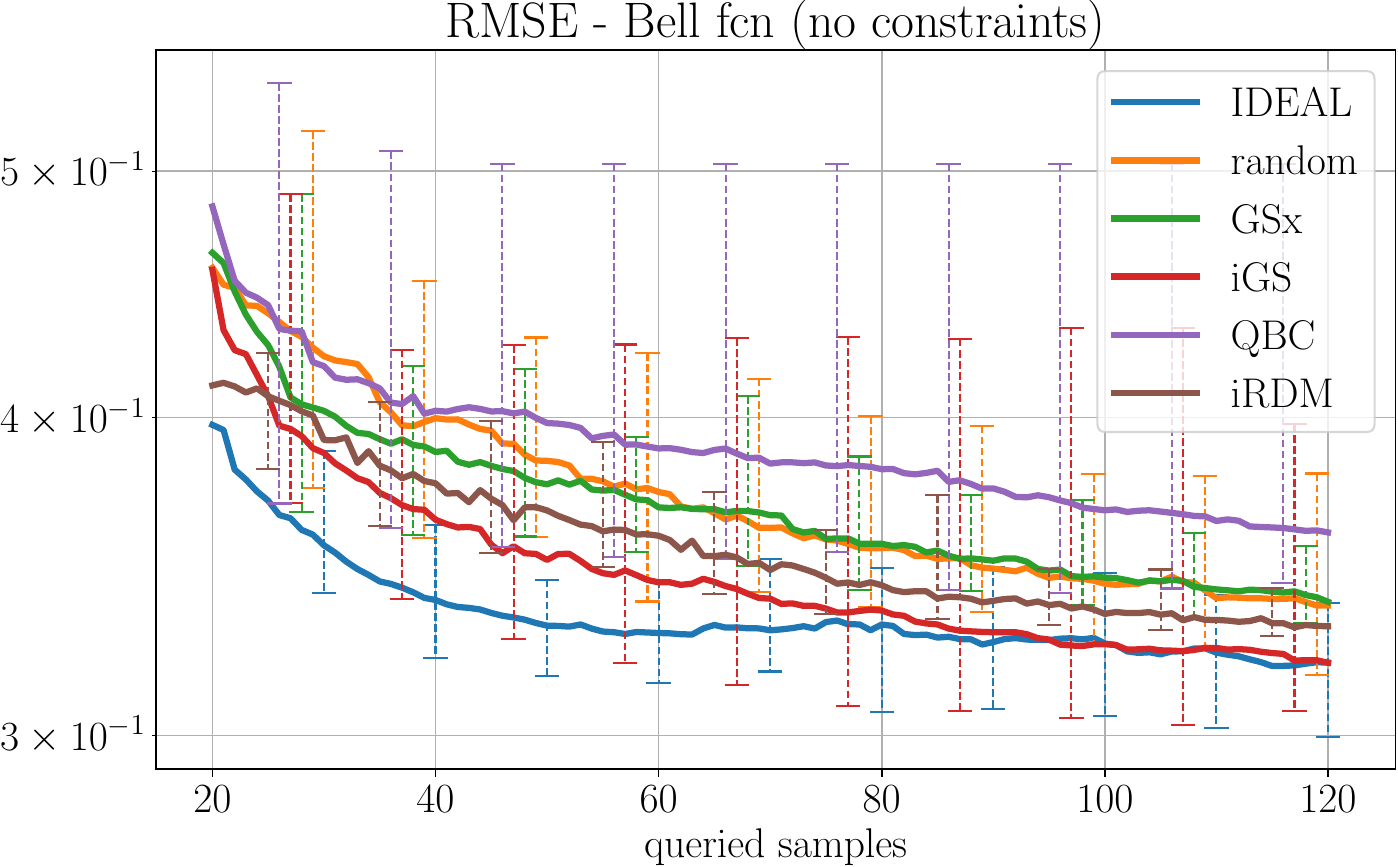}\\[2em]
    \includegraphics[width=0.8\hsize]{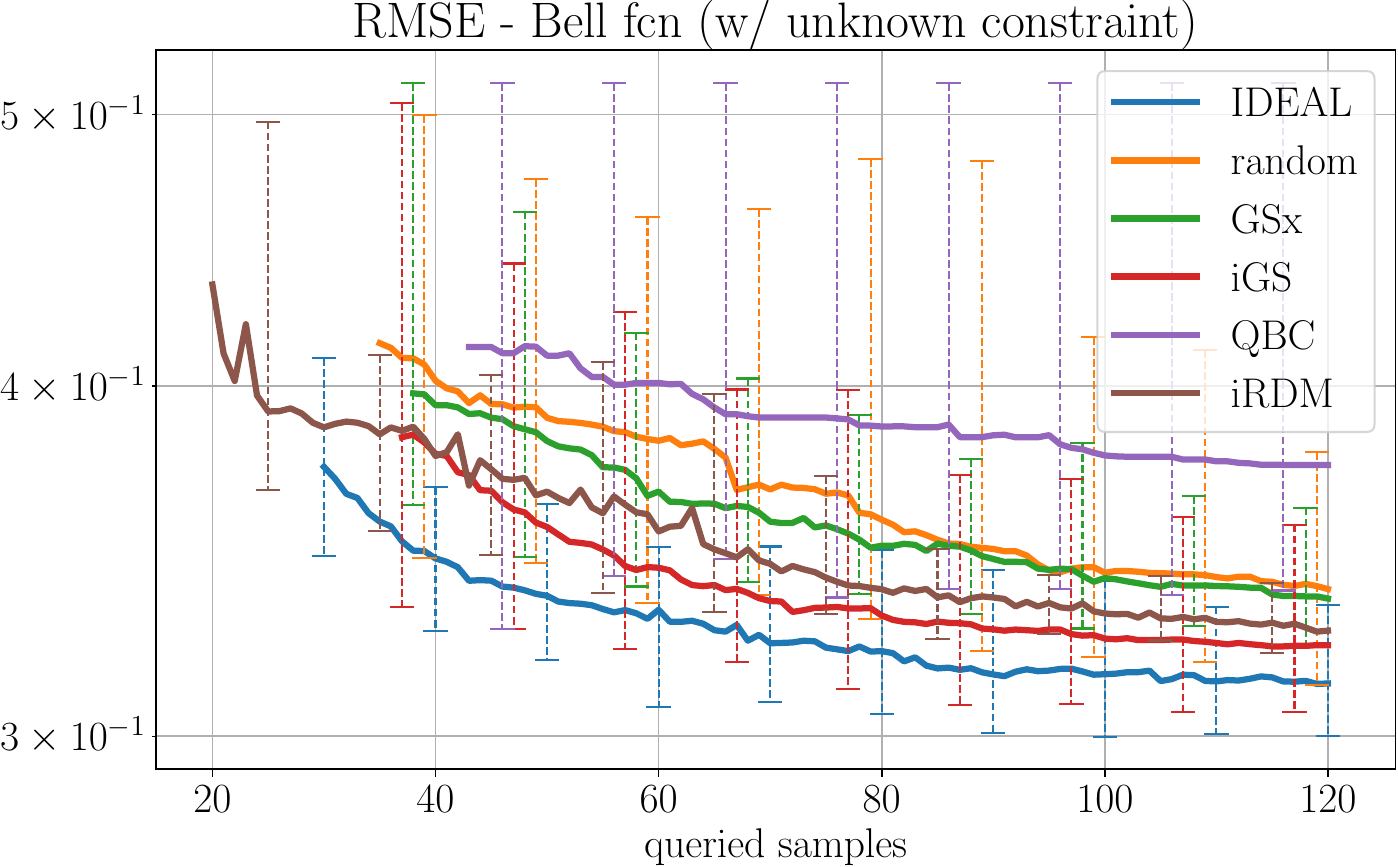}\\
    \end{center}
    \caption{AL problem~\eqref{eq:bell_fcn}, median RMSE
    without (upper plot) and with unknown constraint~\eqref{eq:bell_fnc_unknown_constr}
    (lower plot). Vertical lines denote min and max values}
    \label{fig:bell}
\end{figure*}

In order to test Algorithm~\ref{algo:IDEAL} in the presence of unknown constraints,
we consider data generated by the following bell-shaped function $y:\rr^2\to[0,1]$ 
\begin{equation}
    y(x)=e^{-\left(\left(\frac{3}{2}x_1\right)^2+\left(\frac{3}{2}x_2\right)^2\right)^3}
\label{eq:bell_fcn}
\end{equation}
plotted in Figure~\ref{fig:bell_fcn}.
Algorithm~\ref{algo:IDEAL} is applied with $N_i=10$ and $N_{\rm max}=120$
to fit a nonlinear model via support vector regression with radial basis function (RBF) kernel,
with penalty $\frac{1}{C}=0.1$ for $\ell_2$-regularization and threshold $\epsilon=0.1$. 
Pool-based sampling is used on a set $\XX_P$ 
of $M=1000$ random feature vectors generated uniformly in $[-2,2]\times[-2,2]$.
The median RMSE and its range computed on all vectors $\bar x_i\in\XX_P$
over 50 runs is shown in Figure~\ref{fig:bell} (upper plot). 
A possible reason for the poor performance of {\QBC} in this example
is that the prediction uncertainty estimated by {\QBC} is inaccurate, which 
leads to sampling feature vectors that are in reality not worth sampling. 
In addition, as mentioned earlier, improper sampling leads to poor predictors and hence a poor 
target-uncertainty estimation, so that weak sampling persists. This leads to a waste
of queries.

Next, we add an unknown constraint by only defining $y(x)$ for
$x\in\XX$, where
\begin{equation}
    \XX\eqdef\{x:\ 3x_2\leq\sqrt{3}|x_1|\}
\label{eq:bell_fnc_unknown_constr}
\end{equation}
and repeat the same test,
obtaining the RMSE results shown in Figure~\ref{fig:bell} (lower plot),
where the RMSE is computed only on the feasible vectors $\bar x_i\in\XX_P\cap\XX$.

Note that for {\ideal}, {\random}, {\greedyx}, {\greedyxy}, and {\QBC} 
the RMSE values are not available for $k<N_{\rm init}$. This is due to the fact that, as described in Section~\ref{sec:init}, the
first predictor is trained only after $N_i$ feasible samples have been collected,
which may require $N_{\rm init}>N_i$ queries.
On the contrary, due to its non-incremental nature, when running {\iRDM}
to acquire $k=N_i,N_{i+1},\ldots,N_{\rm max}$ samples, the predictor is always constructed on the available feasible samples, no matter how many feasible samples have been collected
(unless all samples are infeasible, a situation that never occurred in our experiments).

In the above tests, when using {\greedyx} and {\greedyxy} random sampling was employed to 
get the first $N_i$ samples, as in the case of unknown constraints 
the initialization method suggested in~\cite{WLH19} was sometimes failing to get $N_i$ feasible samples within the maximum budget $N_{\rm max}$ of queries.

\begin{figure*}[h!]
    \begin{center}
    \includegraphics[width=0.33\hsize]{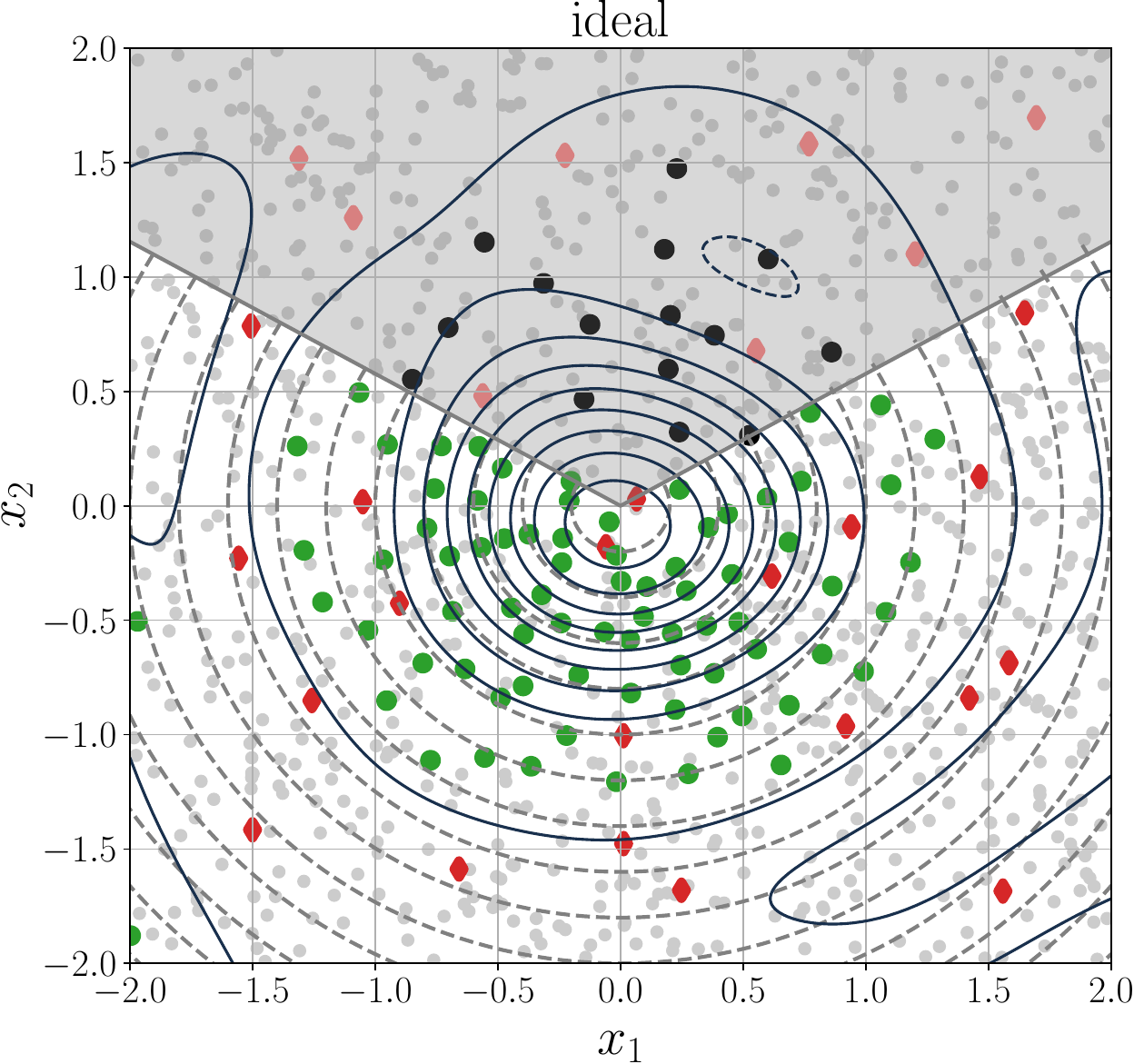}\hspace*{0em}
    \includegraphics[width=0.33\hsize]{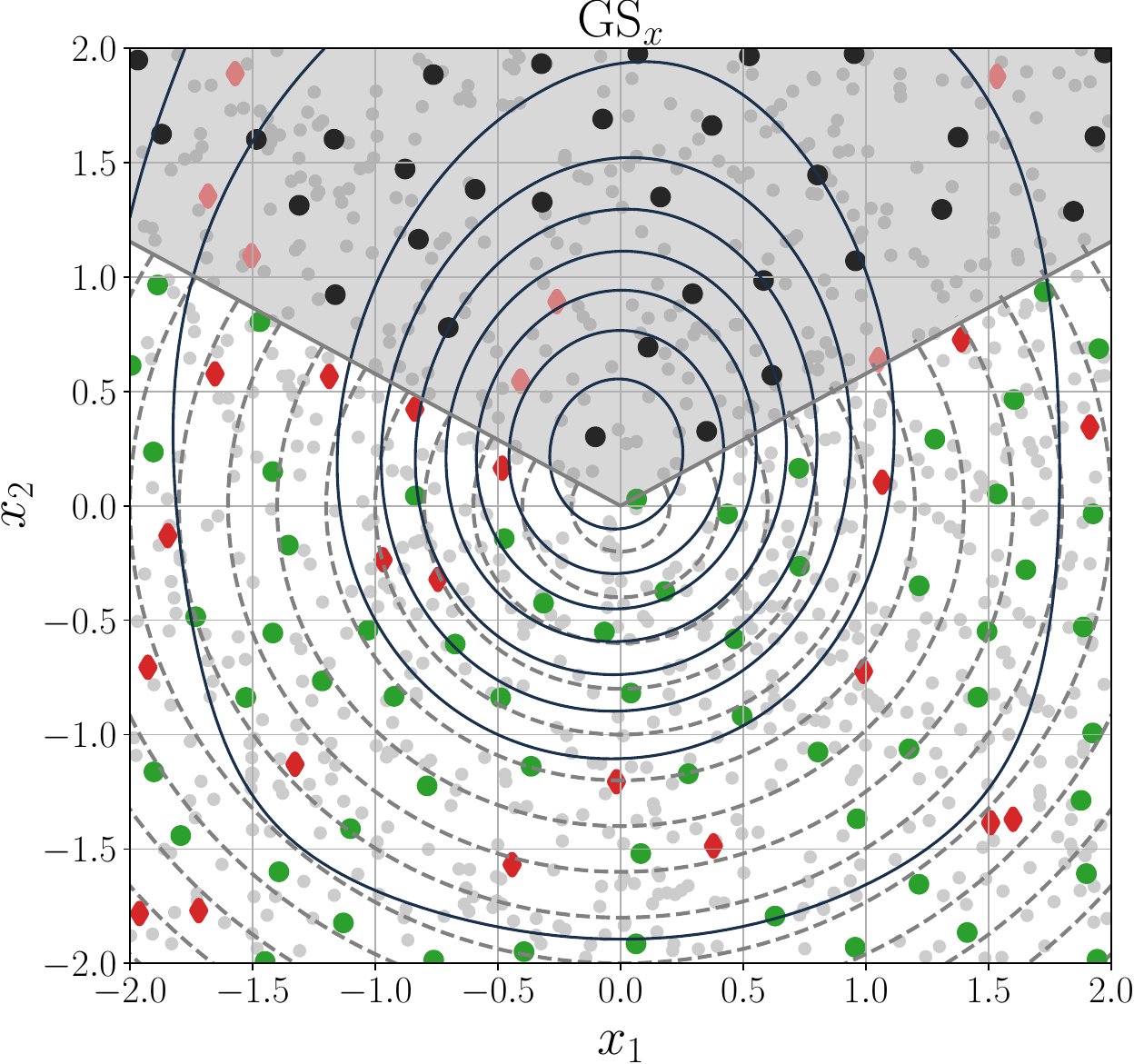}\hspace*{0em}
    \includegraphics[width=0.33\hsize]{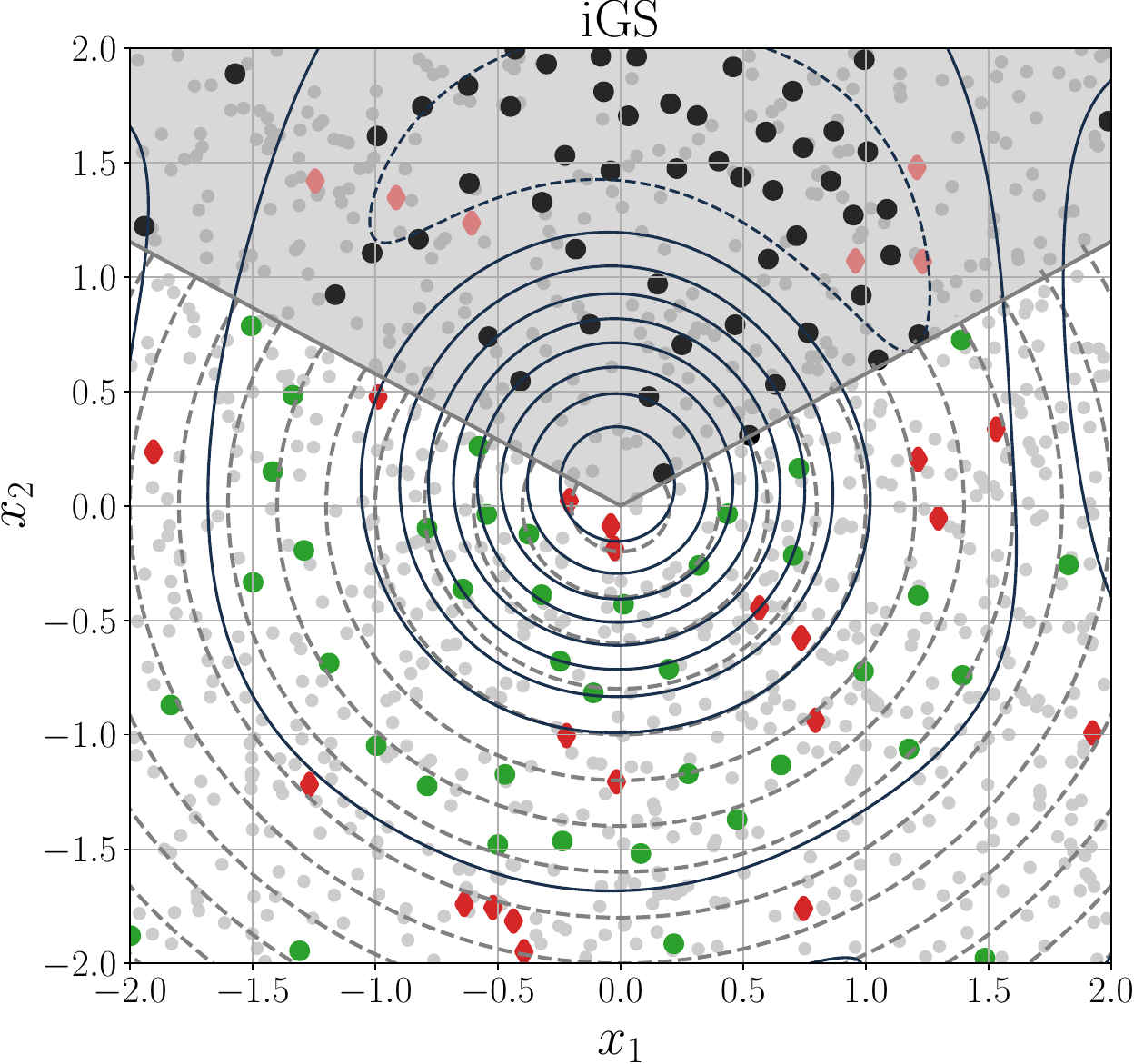}\\[1em]
    \includegraphics[width=0.33\hsize]{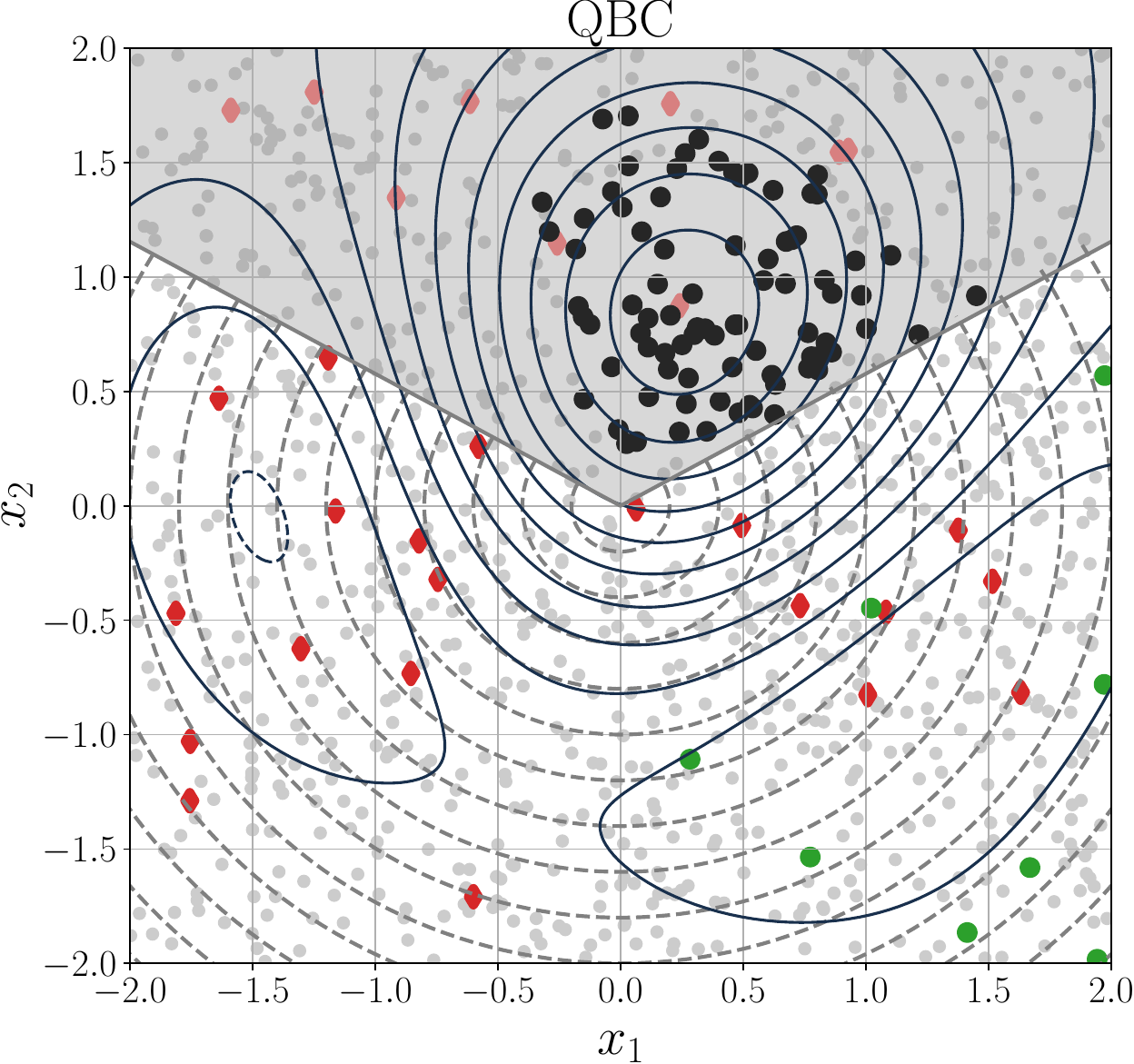}\hspace*{0em}
    \includegraphics[width=0.33\hsize]{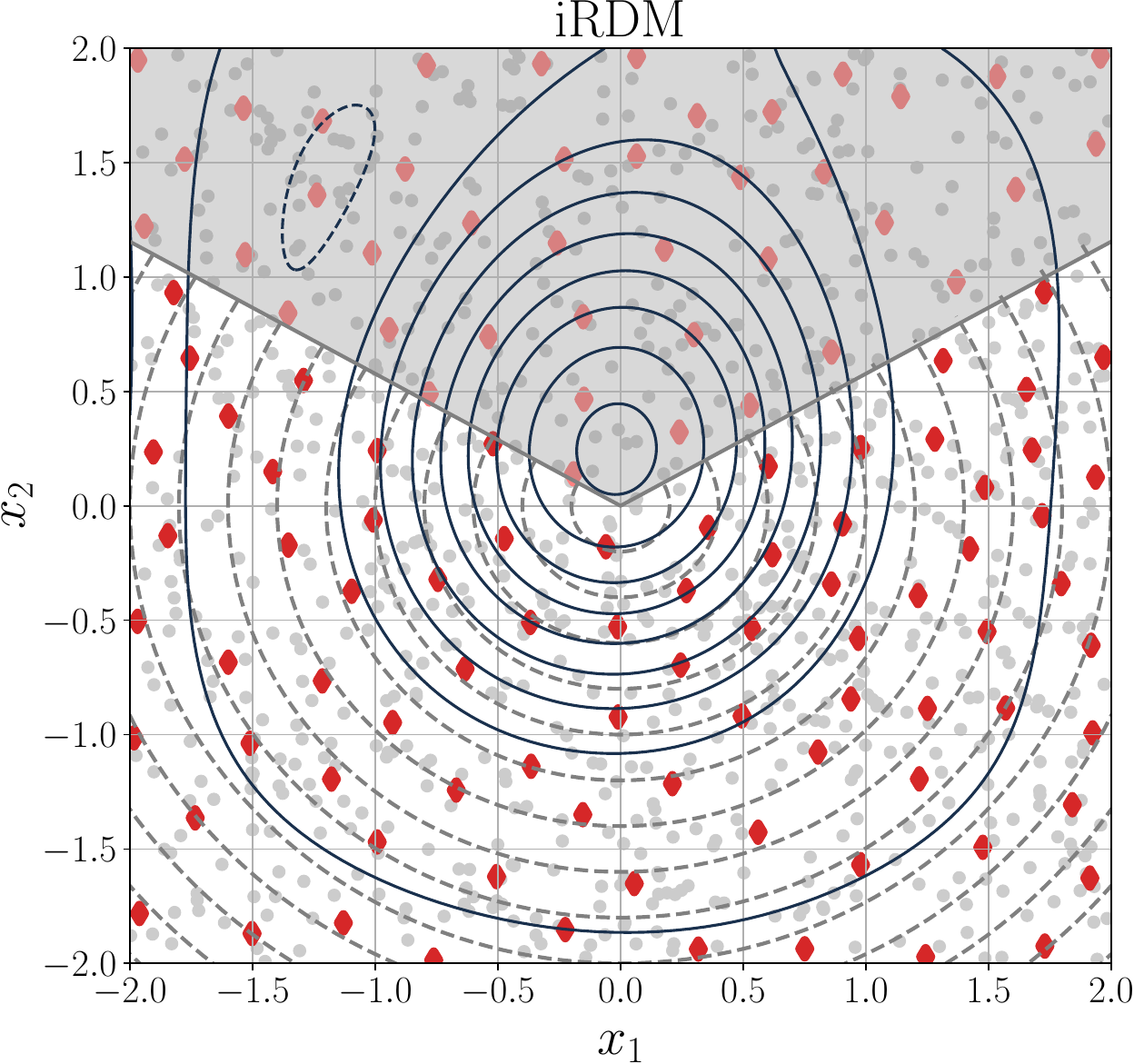}\hspace*{0em}
    \includegraphics[width=0.33\hsize]{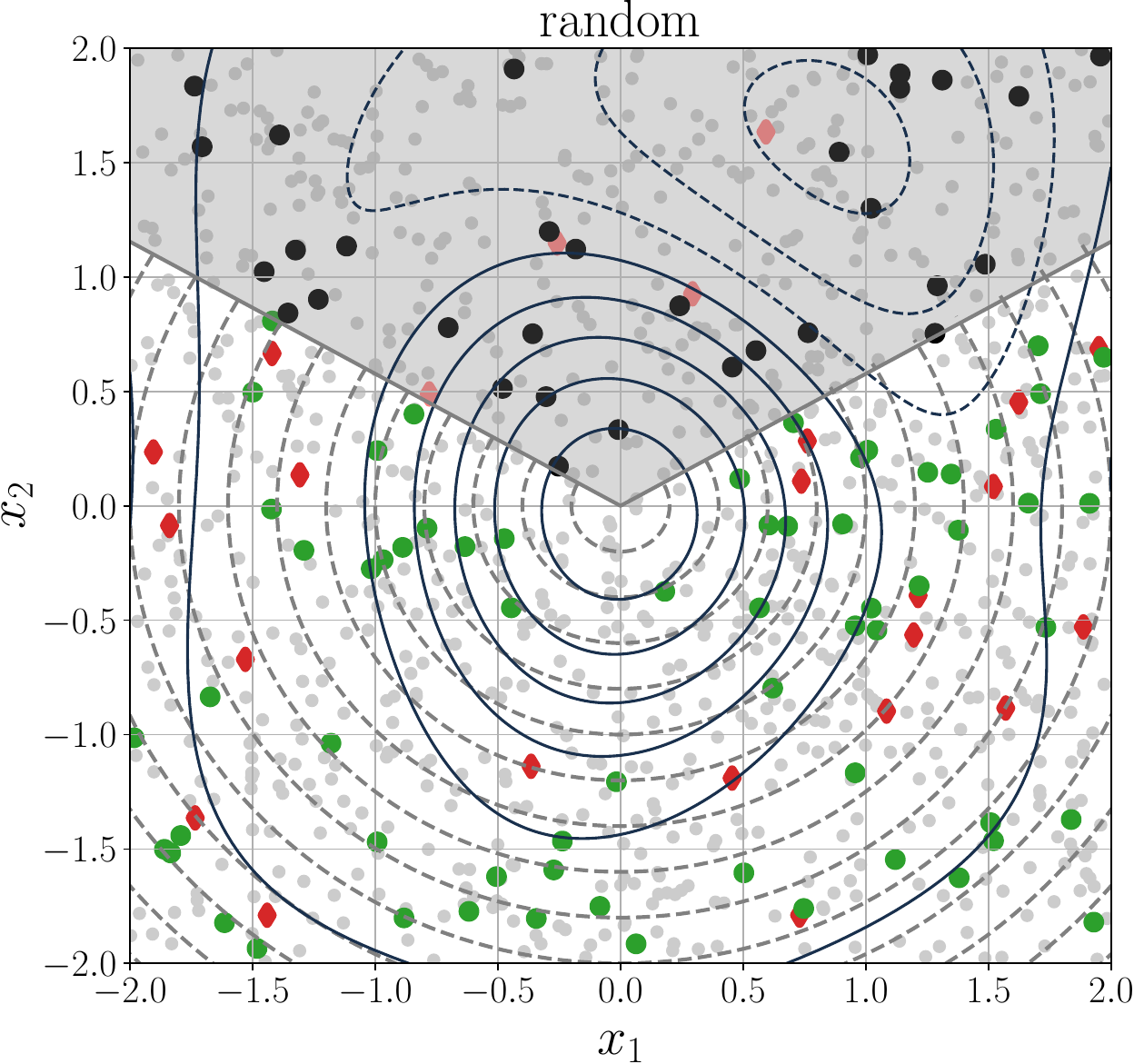}
    \end{center}
    \caption{AL problem~\eqref{eq:bell_fcn} with unknown
    constraint~\eqref{eq:bell_fnc_unknown_constr}: pool $\XX_P$ (gray circles), queried samples (green dots), initial samples (red diamonds), level sets of prediction function (dark blue lines) and
    of the true function~\eqref{eq:bell_fcn} (dashed gray circles)}
    \label{fig:bell-2}
\end{figure*}

Figure~\ref{fig:bell-2} shows the level sets of the
learned classifier $\hat y$ during one of the tests using all the considered methods, the
pool $\XX_P$ of samples (gray circles), the queried samples (green dots), 
the initial samples (red diamonds), and the level sets of the prediction function (dark blue lines)
and of function~\eqref{eq:bell_fcn} (dashed gray circles). It is apparent how {\greedyx} 
and {\iRDM} scatter the points
uniformly no matter whether they are feasible or not, which wastes a large percentage of the queries to get meaningful values $y_i$ (all the $N_{\rm max}$ points acquired by {\iRDM} are marked in red,
as they are selected altogether). 
Similarly, {\greedyxy} also samples infeasible areas of the $x$-space quite consistently,
as it aims at sampling the co-domain of $\hat y$ uniformly
due to the term $d_y$ in~\eqref{eq:d_y}.
Regarding \QBC, it mostly samples the infeasible set, where the $K_{\rm QBC}$ predictors in the committee completely extrapolate due to lack of information and hence tend to disagree 
the most. 
On the other hand,
{\ideal} spontaneously tends to avoid querying infeasible vectors $x\in\XX_P\setminus\XX$
and concentrates most queries where the underlying bell-shaped function has the largest
variations. 

\subsection{Real-world datasets}
We test the proposed AL approach on real-world datasets for regression from the University of California, Irvine (UCI) Machine Learning Repository, Kaggle, and StatLib,
summarized in Table~\ref{tab:UCI-datasets}.

\begin{table}[th]
    \begin{center}
    \begin{tabular}{l|rrr|l|r}
        dataset & $M$ & $n$ & $m$ & $\hat y(x)$ & $N_{\rm max}$\\ \hline
        \texttt{concrete-slump} \tablefootnote{    
            \url{https://archive.ics.uci.edu/ml/datasets/Concrete+Slump+Test}} 
            &103& 7& 1 & neural network & 103\\
            \texttt{auto-mpg} \tablefootnote{\url{https://archive.ics.uci.edu/ml/datasets/auto+mpg}}
            &392& 6 & 1& neural network & 100\\
            \texttt{winequality-white} \tablefootnote{\url{https://archive.ics.uci.edu/ml/datasets/Wine+Quality}} 
            &4898& 7& 1& neural network& 100\\
            \texttt{yacht} \tablefootnote{\url{https://archive.ics.uci.edu/ml/datasets/Yacht+Hydrodynamics}}
            &546& 8& 1 & neural network& 100\\\hline
            \texttt{qsar-aquatic-toxicity} \tablefootnote{\url{https://archive.ics.uci.edu/ml/datasets/QSAR+aquatic+toxicity}}
            &308& 6& 1 & RBF-SVR& 120\\
            \texttt{bodyfat} \tablefootnote{\url{https://www.kaggle.com/fedesoriano/body-fat-prediction-dataset/version/1}}
            &252& 14& 1 & RBF-SVR& 120\\
            \texttt{beer} \tablefootnote{\url{https://www.kaggle.com/datasets/dongeorge/beer-consumption-sao-paulo}}
            &365& 4& 1 & RBF-SVR& 120\\
            \texttt{pm10} \tablefootnote{\url{http://lib.stat.cmu.edu/datasets/PM10.dat}}
            &500& 7& 1 & RBF-SVR& 120\\\hline
    \end{tabular}
    \end{center}
    \caption{Real-world datasets: $M$ = number of available samples in the pool,
    $n$ = number of features, $m$ = 1 (single target), $N_{\rm max}$ = query budget. 
    }
    \label{tab:UCI-datasets}
\end{table}

\begin{figure*}[h!]
    \begin{center}
    \includegraphics[width=.8\hsize]{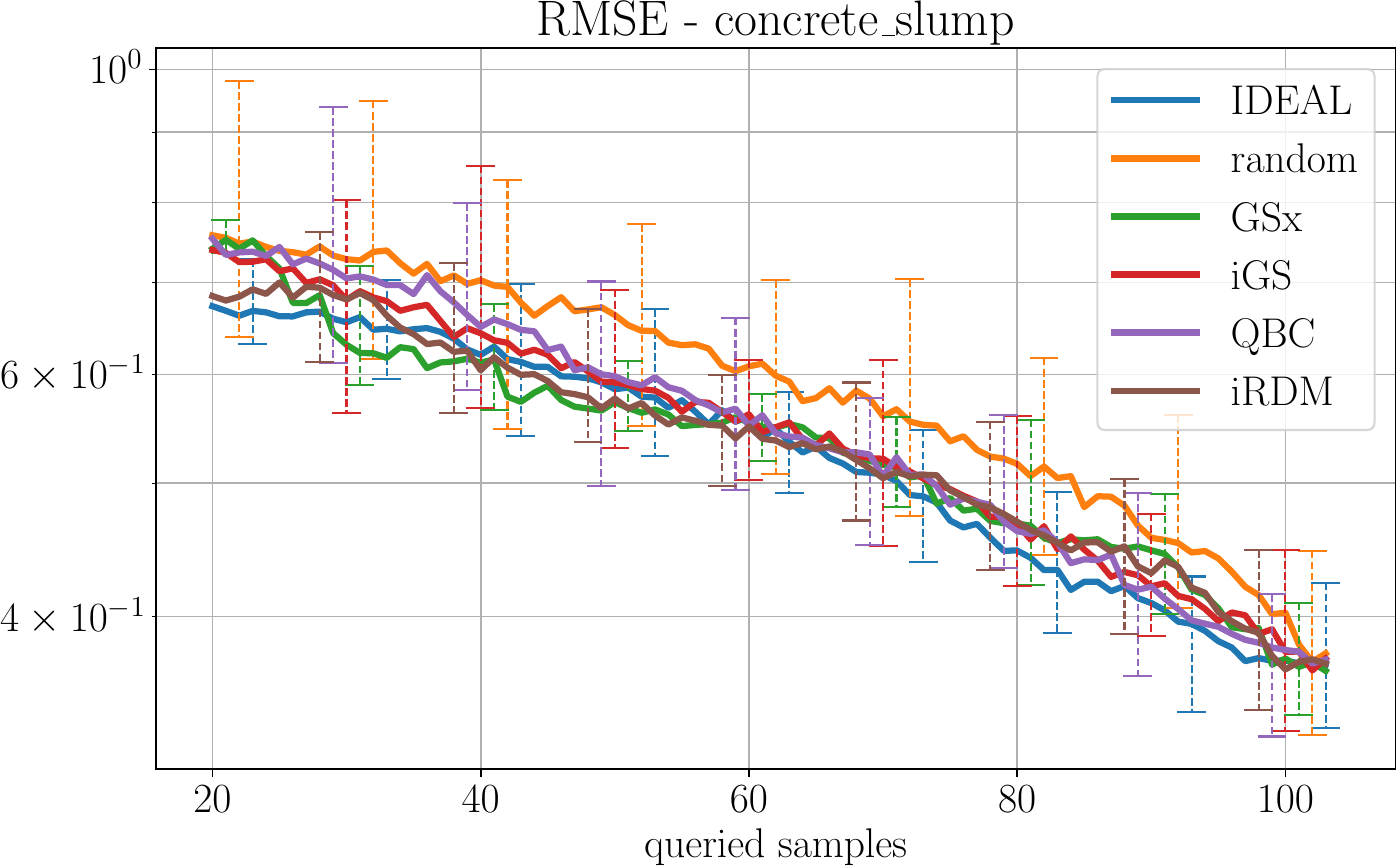}\\[2em]
    \includegraphics[width=.8\hsize]{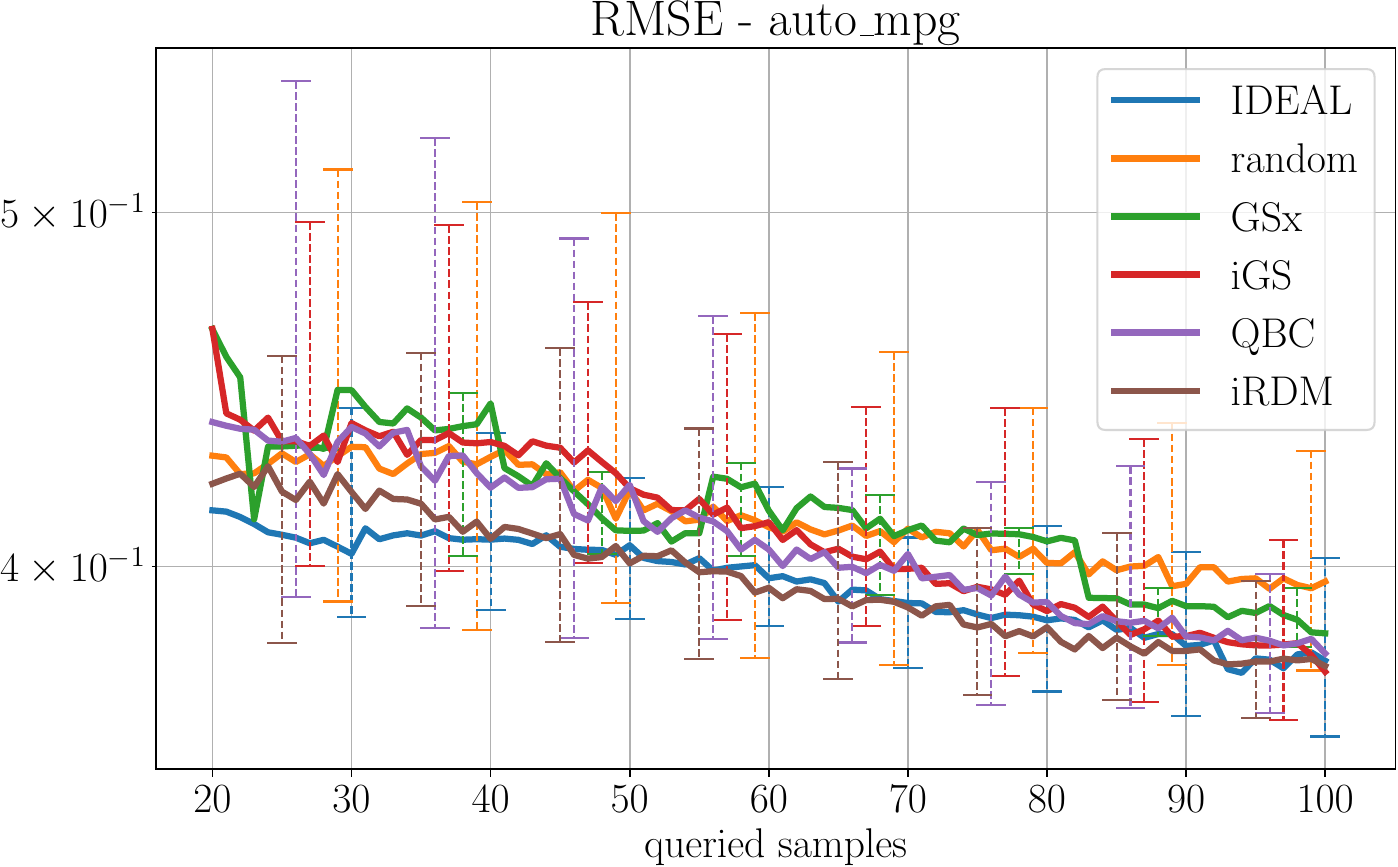}
    \end{center}
    \caption{Regression problems, RMSE results (median and range) on
    \texttt{concrete-slump} (left plot), \texttt{auto-mpg} (right plot) datasets}
    \label{fig:regression1}
\end{figure*}

\begin{figure*}[h!]
    \begin{center}
    \includegraphics[width=.8\hsize]{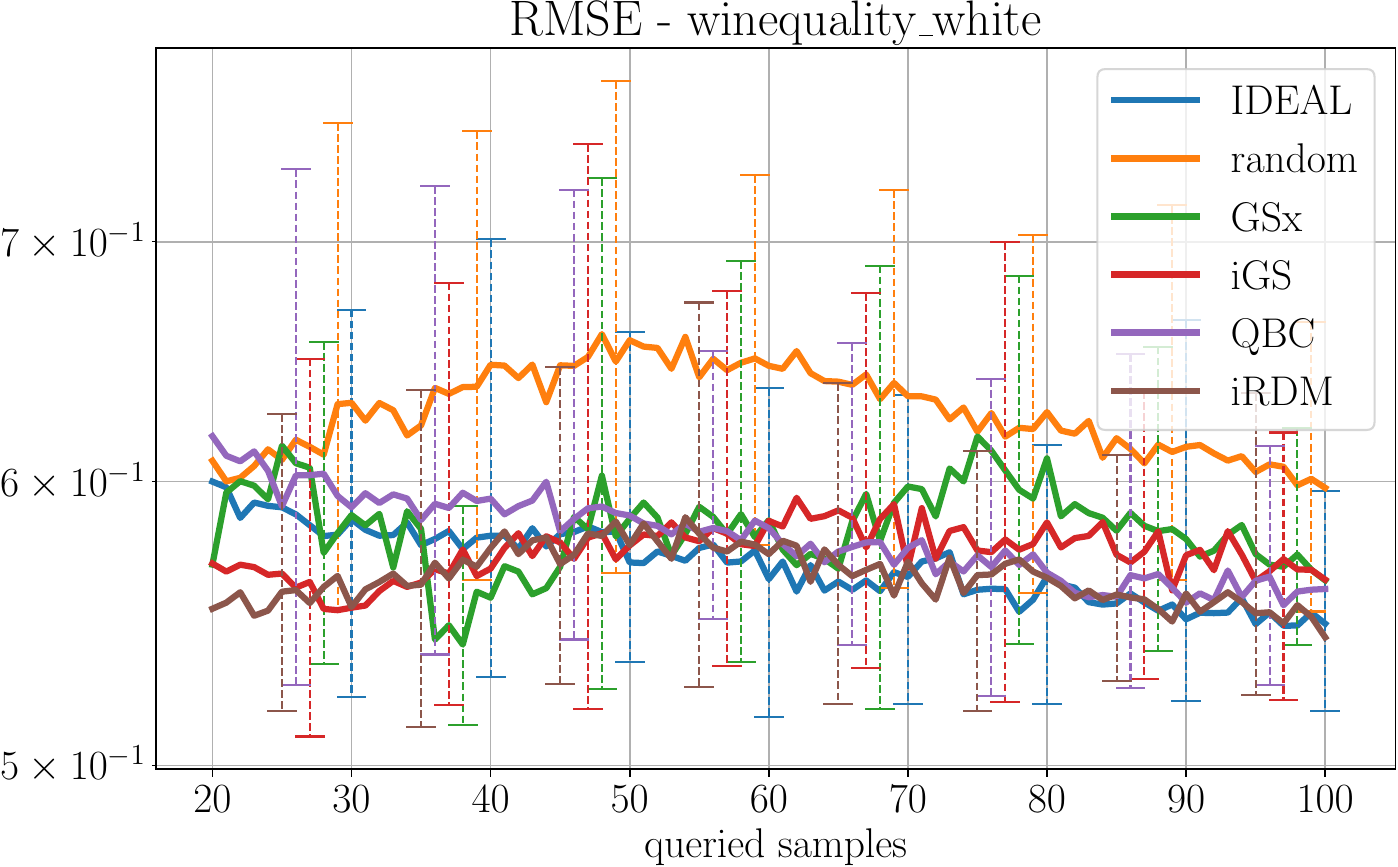}\\[2em]
    \includegraphics[width=.8\hsize]{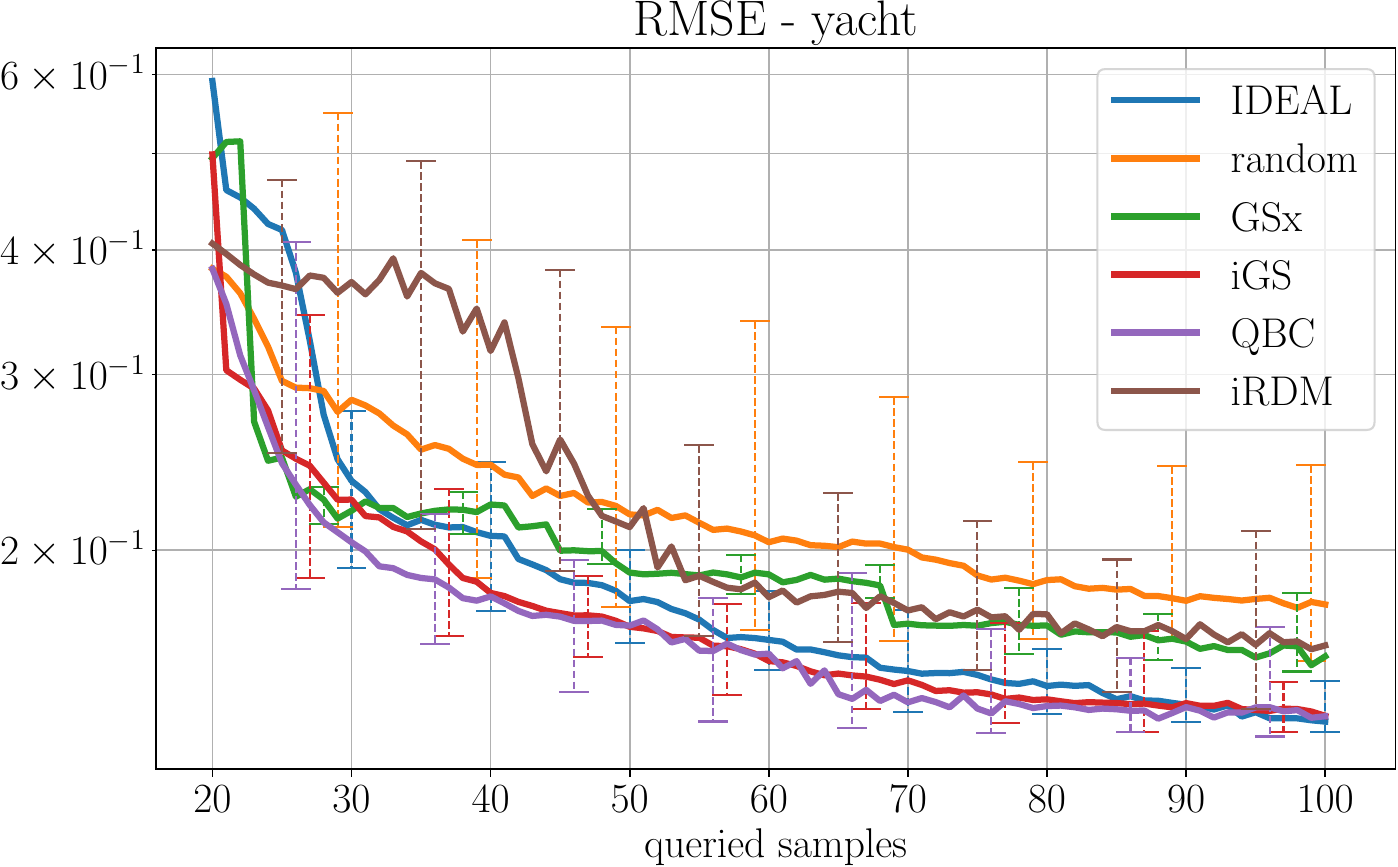}
    \end{center}
    \caption{Regression problems, median RMSE results (median and range) on
    \texttt{winequality-white} (left plot), \texttt{yacht} (right plot) datasets}
    \label{fig:regression2}
\end{figure*}

\begin{figure*}[h!]
    \begin{center}
    \includegraphics[width=.8\hsize]{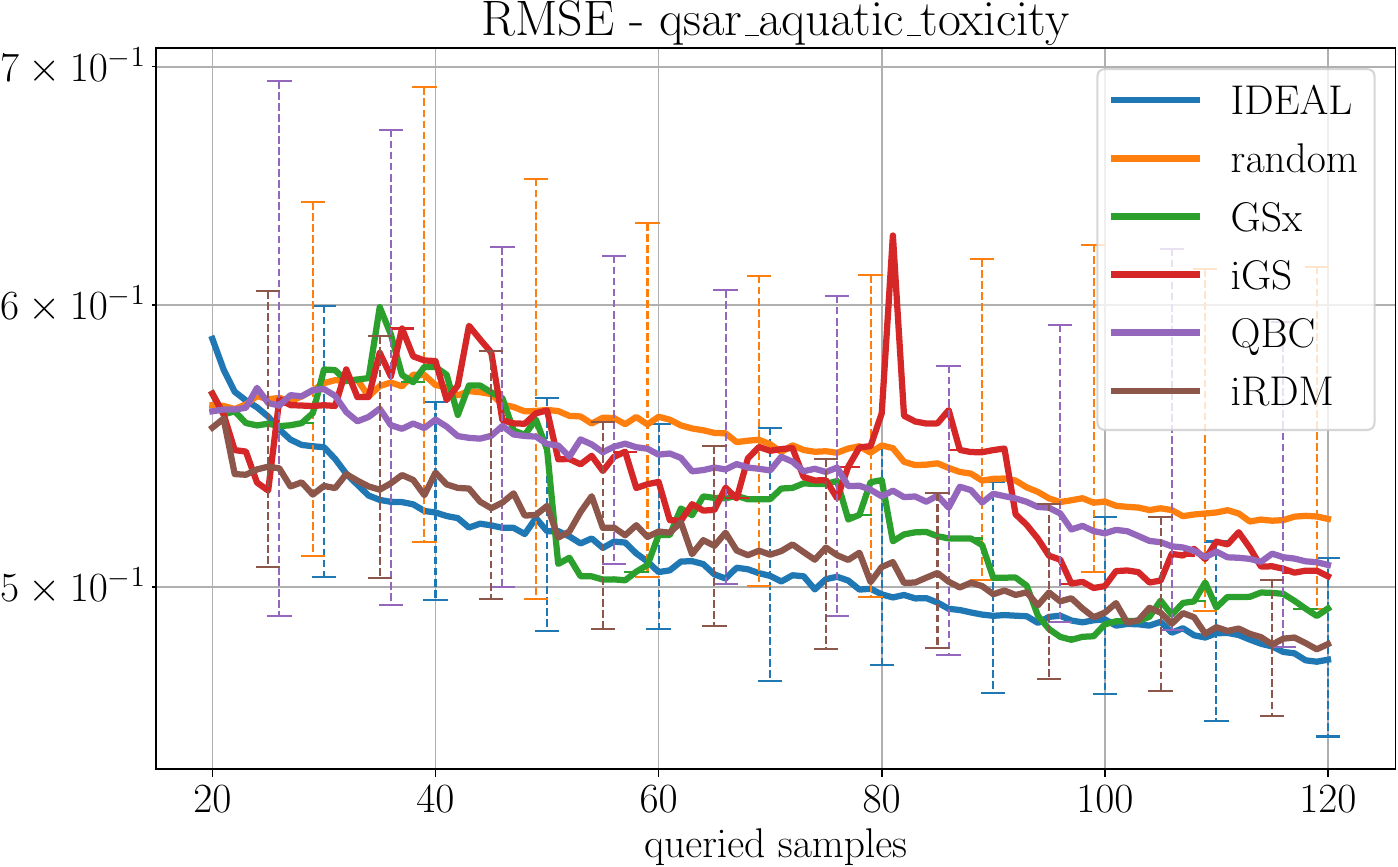}\\[2em]
    \includegraphics[width=.8\hsize]{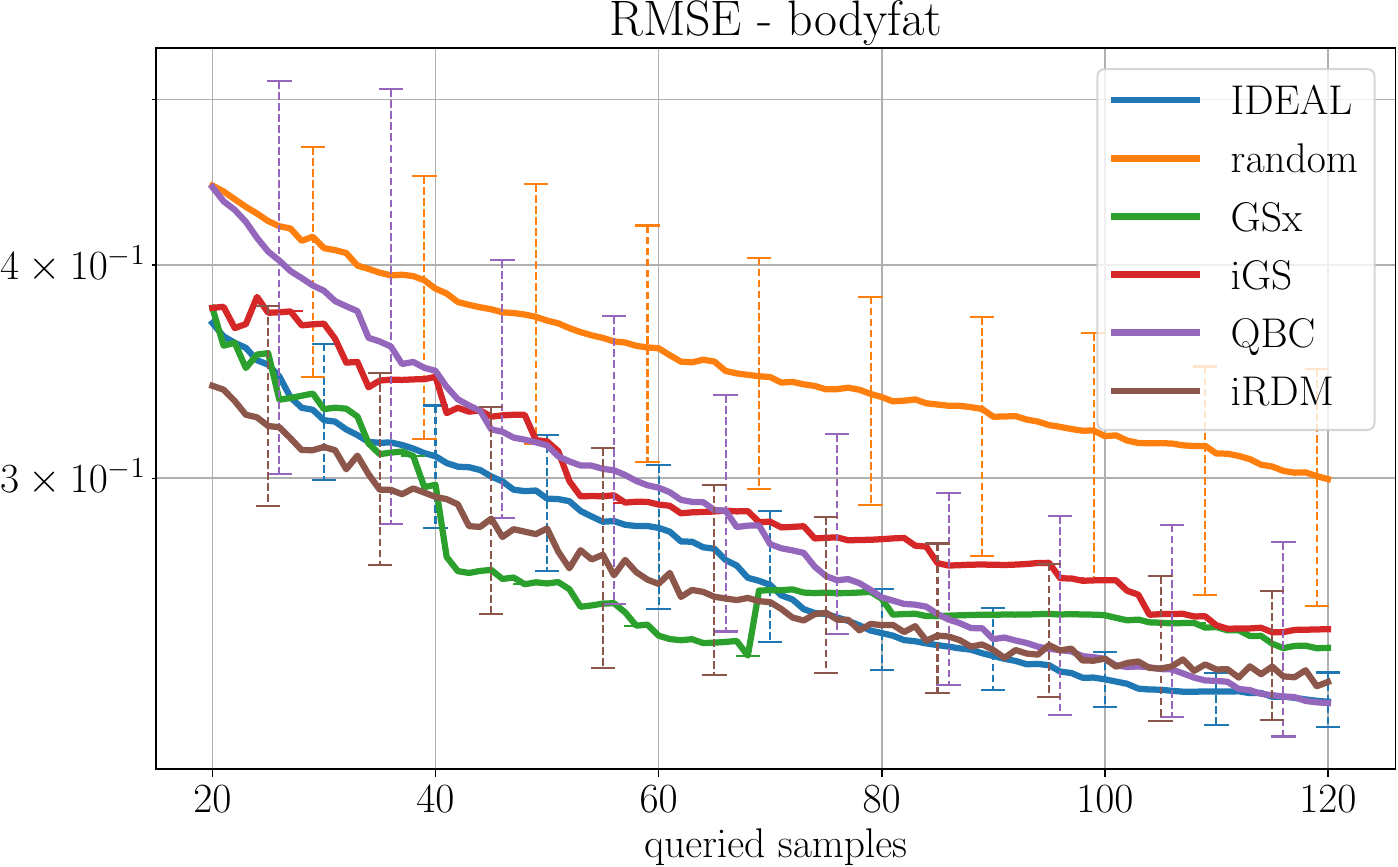}
    \end{center}
    \caption{Regression problems, median RMSE results (median and range) on
    \texttt{qsar-aquatic-toxicity} (left plot), \texttt{bodyfat} (right plot) datasets}
    \label{fig:regression3}
\end{figure*}

\begin{figure*}[h!]
    \begin{center}
    \includegraphics[width=.8\hsize]{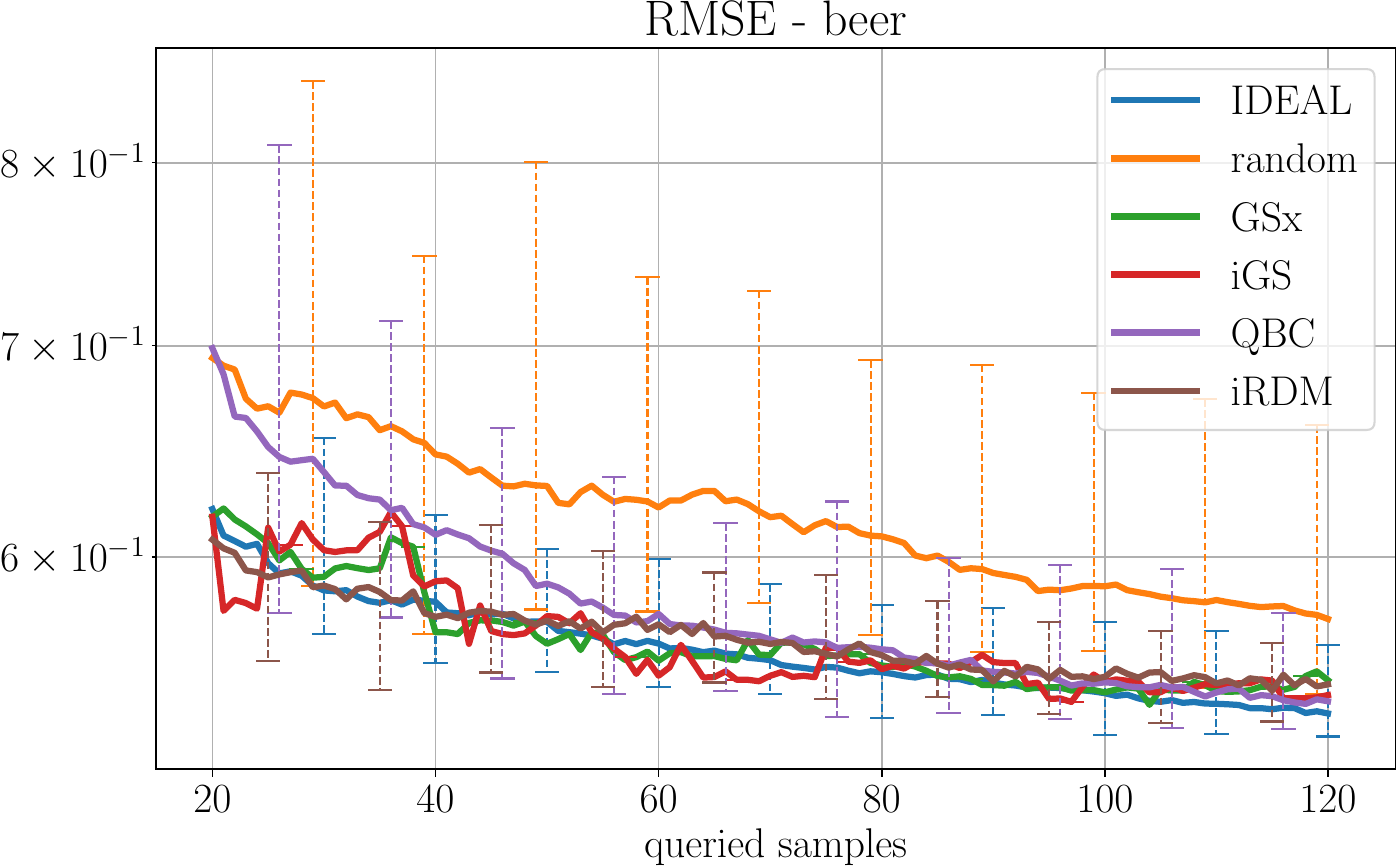}\\[2em]
    \includegraphics[width=.8\hsize]{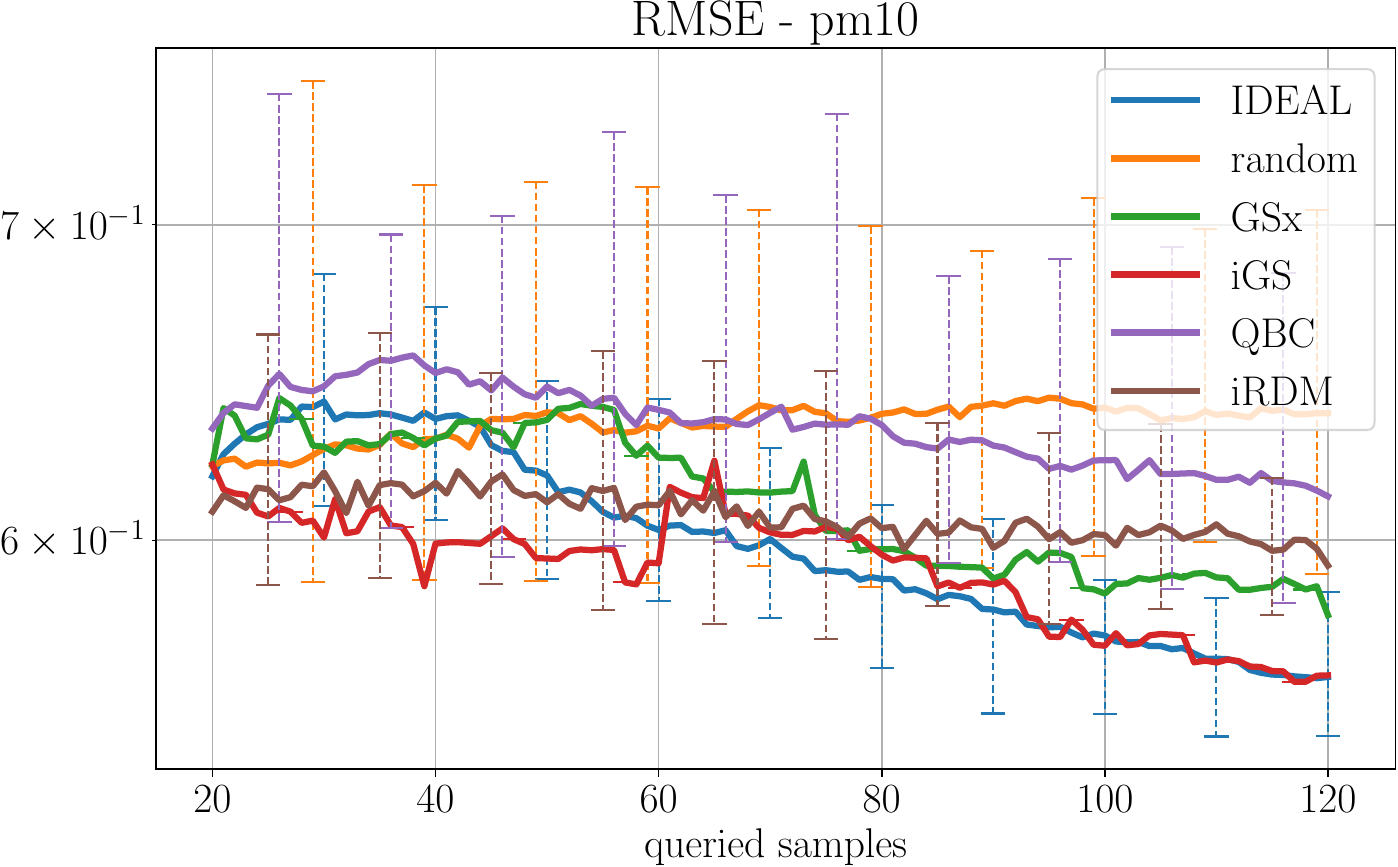}
    \end{center}
    \caption{Regression problems, median RMSE results (median and range) on
    \texttt{beer} (left plot), \texttt{pm10} (right plot) datasets}
    \label{fig:regression4}
\end{figure*}

For the tests described in the upper half of Table~\ref{tab:UCI-datasets},
we train neural networks $\hat y$ with two layers of five neurons
each with logistic activation function and $\ell_2$-regularization term equal to $10^{-2}$ on the vector of weight/bias terms of the model, while for the remaining tests 
we train predictors $\hat y$ using RBF-SVR with penalty $\frac{1}{C}=0.02$ for $\ell_2$-regularization and threshold $\epsilon=0.05$. 
Pool-based AL is used with parameters $N_{\rm init}=20$ and the values of 
$N_{\rm max}$ reported in Table~\ref{tab:UCI-datasets}. Median RMSE results and their ranges over 50 tests are shown in Figures~\ref{fig:regression1}--\ref{fig:regression4}. 

As expected, all the considered AL methods perform better than {\random} when the number
of queries is large enough, with {\QBC} being the method that requires the largest number of
queries to start becoming an effective AL strategy. In spite of its unsupervised AL nature,
{\iRDM} is overall quite effective, sometimes superior to model-based strategies.
In all tests, {\ideal} performs either better or comparably with respect to the other methods,
and is the only method that, at least statistically, seems to perform consistently well with respect to all the considered datasets. The latter feature, consistency, makes it an ``ideal'' candidate 
to face a new active learning problem in practice.

\section{Conclusion}
\label{sec:conclusion}
In this paper we have introduced a new active learning method to solve a 
very broad set of active learning problems of regression. 
The approach is not linked to any particular class of prediction functions 
and supports both pool-based and population-based sampling. The objective
function driving the optimal selection of the next feature vector to query only
requires evaluating the prediction function that has been currently learned
and compare it to the target values acquired so far. This is an advantage compared
to other approaches such as query-by-committee methods in which 
multiple predictors must be trained and evaluated. 

For low-dimensional problems (say up to three features) amenable for population-based
AL, our practical experience is that it is usually more efficient to create a pool $\XX_P$ containing a large but finite set of randomly-selected feature vectors and use pool-based AL instead, i.e., to optimize the sample acquisition problem by enumeration rather than by global optimization over a continuum of values. Our proposed method also seems to be particularly
advantageous to learn functions that have plateaus (this would be the case if applied to classification problems),
because the IDW uncertainty terms tend to be small in regions 
of the feature-vector space where the acquired targets have similar values. 
While this is an advantage, it may also endanger the method, as it may lead to miss
areas of significant change in the underlying function. For this reason, as for global optimization using surrogate functions, we found that a safeguard is to have a large-enough weight $\delta$ on pure exploration, which is entirely independent of the target values acquired and the predictor learned. 

As mentioned at the beginning of Section~\ref{sec:init}, unsupervised AL (such as {\greedyx},
{\iRDM}, K-means, or simply {\random}) is sometimes superior to model-based AL (such as {\ideal}, {\greedyxy}, {\QBC}), see for example Figures~\ref{fig:regression3}--\ref{fig:regression4}.
It would be interesting to investigate the combination of efficient unsupervised AL and model-based 
AL methods, in particular use {\iRDM} to perform the initialization phase of {\ideal}.

We also remark that a rather high variance can be observed when applying all the methods 
considered in our numerical tests. There are several reasons for this. First, when K-means
is applied for initialization, the final cluster centroids found may depend heavily on their
initial values, due to the fact that K-means is a coordinate-descent method that 
is not guaranteed to reach a global minimum.
Moreover, in the case of active learning of neural network models, additional variance
is due to the non-convexity of the learning problem, which may lead to largely different
prediction models depending on the random initial values of the trained weights/bias terms.
Further variance is suffered by {\QBC} due to the random generation of bootstrap samples.

Future research will be devoted to analyze in depth the use of {\ideal} to solve classification problems, to adapt the weight on the exploration term $\delta$ automatically while learning, to extend the method to streaming data to support online learning problems, and to active learning for
identification of dynamical systems.


\end{document}